\documentclass[acmtog]{acmart}
 
\usepackage{xspace}

\usepackage{booktabs}
\usepackage{multirow}
\usepackage{graphicx}
\usepackage{pifont}

\newcommand{\eg}{\textit{e.g.}\xspace}
\newcommand{\methodname}{\textbf{PAct}\xspace}

\AtBeginDocument{%
  }

\copyrightyear{2026}
\acmYear{2026}
\setcopyright{cc}
\setcctype{by-nc-nd}
\acmConference[SA Conference Papers '26]{SIGGRAPH Asia 2026 Conference Papers}{December 01--04, 2026}{Kuala Lumpur, Malaysia}
\acmBooktitle{SIGGRAPH Asia 2026 Conference Papers (SA Conference Papers '26), December 01--04, 2026, Kuala Lumpur, Malaysia}
\acmDOI{10.1145/3829340.3842230}
\acmISBN{979-8-4007-2842-6/2026/12}

\acmSubmissionID{1488}

\begin{document}

\title{\methodname: Part-Decomposed Single-View Articulated Object Generation}

\author{Qingming Liu}
\email{qingmingliu@foxmail.com}
\affiliation{%
  \institution{The Chinese University of Hong Kong, Shenzhen}
  \city{Shenzhen}
  \country{China}
}
\affiliation{%
  \institution{DexForce Technology}
  \city{Shenzhen}
  \country{China}
}
\author{Xinyue Yao}
\email{xinyueyao1@link.cuhk.edu.cn}
\affiliation{%
  \institution{The Chinese University of Hong Kong, Shenzhen}
  \city{Shenzhen}
  \country{China}
}

\author{Shuyuan Zhang}
\affiliation{%
  \institution{The Chinese University of Hong Kong, Shenzhen}
  \city{Shenzhen}
  \country{China}
}
\email{rayzhang707@gmail.com}

\author{Yueci Deng}
\email{yuecideng@link.cuhk.edu.cn}
\affiliation{%
  \institution{The Chinese University of Hong Kong, Shenzhen}
  \city{Shenzhen}
  \country{China}
}
\affiliation{%
  \institution{DexForce Technology}
  \city{Shenzhen}
  \country{China}
}

\author{Guiliang Liu}
\affiliation{%
  \institution{The Chinese University of Hong Kong, Shenzhen}
  \city{Shenzhen}
  \country{China}
}
\email{guiliangliu@cuhk.edu.cn}

\author{Zhen Liu}
\authornote{Corresponding author.}
\affiliation{%
  \institution{The Chinese University of Hong Kong, Shenzhen}
  \city{Shenzhen}
  \country{China}
}
\email{zhenliu@cuhk.edu.cn}

\author{Kui Jia}
\affiliation{%
  \institution{The Chinese University of Hong Kong,Shenzhen}
  \city{Shenzhen}
  \country{China}
}
\affiliation{%
  \institution{DexForce Technology}
  \city{Shenzhen}
  \country{China}
}
\email{kuijia@cuhk.edu.cn}

\renewcommand{\shortauthors}{Liu et al.}

\begin{teaserfigure}
\centering
    \includegraphics[width=0.99\textwidth,trim=0 160bp 0 110bp,clip]{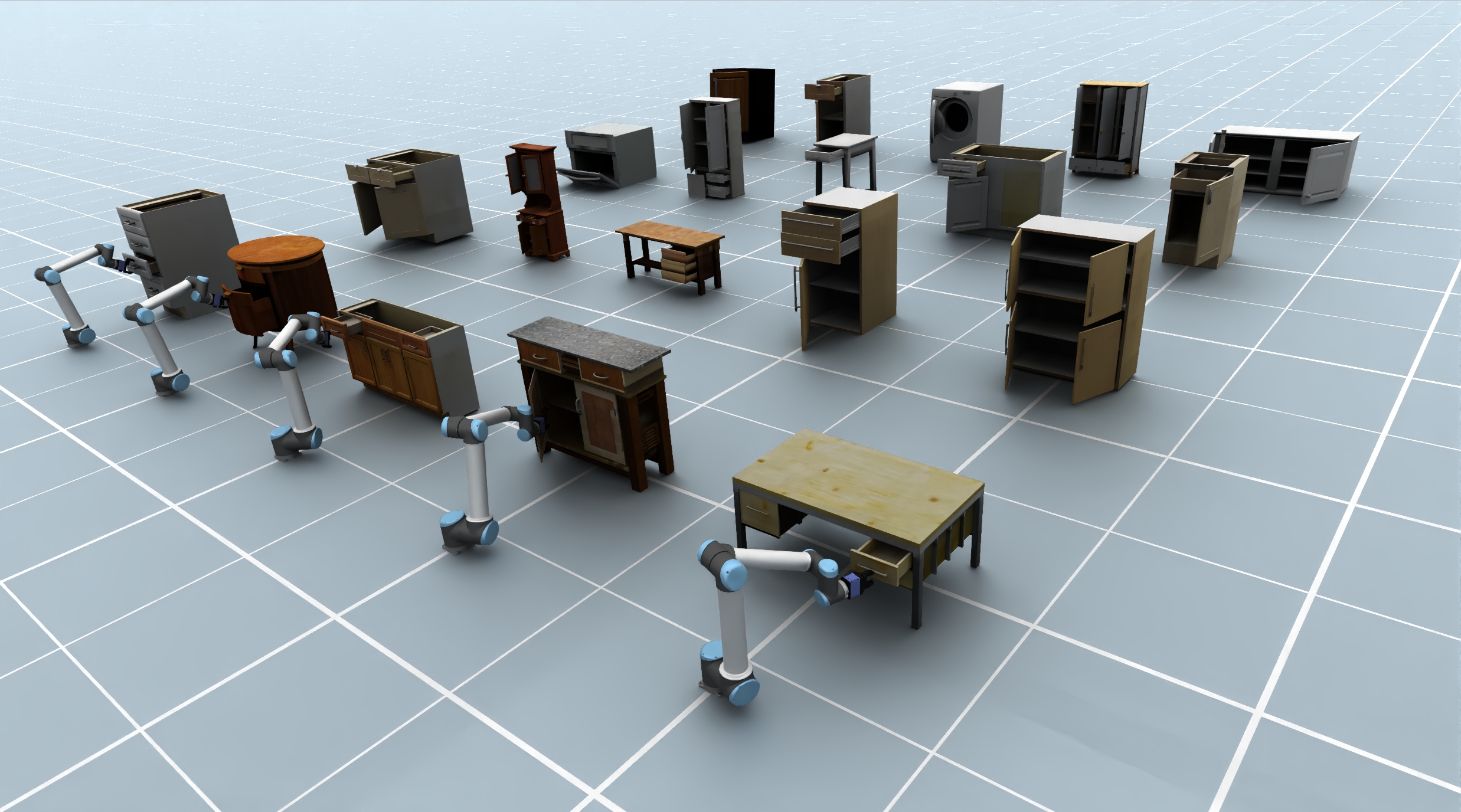}
    \caption{
    Given a single image, \methodname generates an articulated 3D object by predicting a part-decomposed structure, synthesizing high-fidelity part geometry and appearance, and estimating articulation parameters  for downstream applications such as building simulation environments for embodied intelligence.
    } %
\Description{A simulated environment containing articulated 3D furniture and appliances generated by PAct, including cabinets, desks, tables, storage units, and a washing machine, with doors and drawers shown in different open states. Several robotic arms interact with the generated objects, illustrating their use as articulated assets for embodied-AI simulation.}
\end{teaserfigure}

\begin{abstract}

Articulated 3D assets are central to interactive 3D applications, including embodied AI, robotics, and VR/AR, where functional part decomposition and valid kinematic motion are essential. Yet producing high-fidelity articulated assets remains difficult to scale, as it requires reliable part decomposition and kinematic rigging. Existing approaches largely fall into two paradigms: optimization-based reconstruction or score-distillation methods, which can be accurate but often require tens of minutes to hours per instance, and feedforward methods that rely on template or part retrieval, producing plausible results that may not match the specific structure and appearance of the input observation.
We present \methodname, a feedforward, part-centric framework that converts a single-view input into an articulated 3D asset. Built on a pretrained 3D-native generative model, it uses part-decomposed latents and explicit part signals for controllable part decomposition, while alternating global and within-part attention to preserve both part-specific structure and cross-part coherence. It further predicts articulation parameters using multi-step denoising features, effectively leveraging the pretrained 3D generative prior for kinematic estimation.
Experiments on common articulated-object categories demonstrate that \methodname consistently improves part geometry, articulation plausibility, and input-image consistency over optimization-based and feedforward baselines, while substantially reducing inference time.
Project page: \href{https://PAct-project.github.io}{\color{magenta}{https://PAct-project.github.io} }

\end{abstract}

\ccsdesc[500]{Computing methodologies~Vision for robotics}
\ccsdesc[300]{Computing methodologies~Motion processing}
\ccsdesc[500]{Computing methodologies~Reconstruction}
\keywords{Generative models, Articulated objects, 3D asset generation}

\maketitle

\section{Introduction}

Articulated objects are pervasive in everyday environments. Faithfully replicating our interactive physical world for applications such as virtual reality and embodied-AI simulation requires reconstructing not only object geometry and appearance, but also functional part decomposition and articulation, ideally from single-view images. Despite recent progress in 3D reconstruction and generative 3D modeling, recovering object-specific articulation parameters together with high-quality 3D representations from visual inputs remains challenging. In practice, reconstructed assets often require extensive manual cleanup, or even modeling from scratch, before they are usable in production settings.

Existing efforts to create articulated objects from visual observations can be roughly grouped into two paradigms: methods based on iterative optimization and feedforward methods. The former reconstruct articulated 3D assets by fitting geometry and articulation parameters with explicit visual supervision, such as images or videos~\cite{liu2025building, liu2025videoartgs,jiayi2023paris,hsu2023ditto,jiang2022ditto}, and/or by leveraging signals from pretrained generative models through techniques such as score distillation sampling~\cite{chen2025freeart3d,qiu2025articulate}. Although these approaches can be accurate when rich and well-suited multi-view supervision is available, they are costly, often requiring minutes per instance, and can be prone to optimization failures because the underlying reconstruction problem is highly ill-posed. This issue becomes more severe in the single-view setting, where visual constraints are weaker and the result depends more heavily on generative priors and optimization stability. In contrast, feedforward reconstruction methods train networks that directly map visual inputs to reconstructed 3D assets~\cite{wang2024pf,hong2023lrm,tang2024lgm,szymanowicz2024splatter,xiang2024trellis,xiang2025trellis2,yang2024hunyuan3d}. Recent works further extend this paradigm to simulation-ready articulated objects by inferring part structures, such as bounding boxes, together with articulation relations~\cite{yuan2025larmlargearticulatedobjectreconstruction, liu2024singapo, wu2025dipo}. However, existing feedforward methods often rely on template-based retrieval, which rarely matches the target instance accurately, or train reconstruction models from scratch, which can generalize poorly under limited articulated-object training data.

Motivated by these challenges, we propose \methodname, a feedforward, part-decomposed generative framework for articulated object creation from a single image. We model an object as a set of movable parts, each with 3D geometry and appearance, together with articulation attributes (\textit{e.g.}, semantic label, joint type, axis, pivot, and motion range). To achieve generalizable and faithful reconstruction, \methodname is built by finetuning TRELLIS~\cite{xiang2024trellis}, a pretrained two-stage 3D-native generative model, to directly generate part-decomposed 3D assets and their articulation parameters. Specifically, each object is encoded as a set of part latents, and the denoising transformer augments global attention with within-part attention to refine individual parts while preserving cross-part consistency. Along the diffusion sampling trajectory, we extract intermediate denoising features at multiple timesteps, aggregate them per part, and feed the resulting part descriptors into a lightweight MLP to predict articulation parameters. This unified design enables fast feedforward inference, remains faithful to the observed instance, and supports controllable part decomposition, assembly, and articulation in a single generation pipeline.

\section{Related Works}

\paragraph{Articulated Object Reconstruction.}
Articulated object reconstruction aims to recover both part geometry/appearance and kinematics from observations.
Early efforts leverage interaction or multi-state observations to build digital twins and estimate joint parameters, e.g., Ditto learns part-level geometry and articulation from before/after interaction pairs~\cite{jiang2022ditto, hsu2023ditto}, while SfA~\cite{nie2022structure} explores interaction-driven structure discovery for unseen objects.
Recent works increasingly exploit neural implicit fields and differentiable rendering to jointly optimize shape and motion: PARIS~\cite{jiayi2023paris} reconstructs part-level implicit fields and motion parameters in a self-supervised manner from multi-view observations of different states, and REACTO~\cite{song2024reacto} extends reconstruction to casually captured monocular videos with quasi-rigid deformation.
Alongside method advances, large-scale real data and benchmarks facilitate evaluation in the wild, such as MultiScan~\cite{mao2022multiscan} for RGBD scanning and annotation of articulated objects in indoor scenes.
More recently, 3DGS has become a strong representation for photorealistic articulated replicas: ArtGS~\cite{liu2025building} and VideoArtGS~\cite{liu2025videoartgs} align information across object states, and SplArt~\cite{lin2025splart} further estimates kinematics and part-level reconstruction with self-supervised 3DGS~\cite{kerbl3Dgaussians} training from multi-state images.
Finally, a growing line of work targets \emph{feed-forward} articulated reconstruction
for efficiency.
Real2Code predicts articulation via code generation from visual cues~\cite{mandi2024real2code};
URDFormer predicts articulated scene structures from RGB images for simulation~\cite{chen2024urdformer}.
LARM reconstructs articulated objects from sparse views in two states~\cite{yuan2025larmlargearticulatedobjectreconstruction}, while ART jointly predicts geometry, texture, and articulation from multi-state images~\cite{li2025art}.

\paragraph{Articulated Object Generation.}
NAP and CAGE model articulated objects through part structures and motion attributes~\cite{lei2023nap,liu2024cage}, while SINGAPO and DIPO condition generation on single-image and dual-state observations, respectively~\cite{liu2024singapo,wu2025dipo}.
MeshArt autoregressively generates articulation-aware structures and part meshes~\cite{gao2024meshart}.
VLM-assisted approaches include Articulate-Anything for mesh retrieval and articulation-code generation~\cite{le2024articulate}, Articulate-AnyMesh for articulating rigid meshes~\cite{qiu2025articulate}, and Kinematify for high-DoF topology inference and joint optimization~\cite{wang2025kinematif}.
SPARK combines VLM-inferred structures, part generation, and joint refinement~\cite{he2025spark}.
FreeArt3D uses a pretrained 3D diffusion prior for training-free optimization from multi-state images~\cite{chen2025freeart3d}, whereas DreamArt combines single-image 3D generation, amodal part completion, and motion optimization guided by a video diffusion model~\cite{lu2025dreamart}.
ArtiLatent and UniArt jointly encode geometry and articulation in unified latent~\cite{chen2025artilatent,jin2025uniart}.
For scalable asset creation, Infinite Mobility uses procedural generation~\cite{lian2025infinite}, while Articraft uses LLM agent and automated validation to build Articraft-10K~\cite{zhou2026articraft}.
Our method adapts pretrained 3D priors through mask-conditioned local--global denoising and predicts articulation from aggregated multi-step denoiser features.

\paragraph{3D Generative Model.}

Tripo series enables fast and high-quality generation~\cite{TripoSR2024,li2025triposg}.
TRELLIS introduces structured latents supporting multiple 3D output formats~\cite{xiang2024trellis}, while TRELLIS.2 uses compact native 3D latents to encode geometry and material attributes~\cite{xiang2025trellis2}.
The Hunyuan3D series advances textured 3D asset generation, with PBR support in version 2.5~\cite{hunyuan3d22025tencent,lai2025hunyuan3d25highfidelity3d,yang2024hunyuan3d}.
Seed3D targets simulation-ready assets and physically based materials~\cite{seed3D1.0}.
Beyond holistic assets, PartCrafter jointly generates part meshes through compositional latent diffusion~\cite{lin2025partcrafter}, while PartPacker organizes variable numbers of parts into two complementary volumes~\cite{tang2024partpacker}.
OmniPart uses 2D masks to guide layout planning, followed by simultaneous part synthesis~\cite{yang2025omnipart}.
PartGen reconstructs complete parts through multi-view segmentation and completion~\cite{chen2024partgen}, whereas AutoPartGen generates parts autoregressively~\cite{chen2025autopartgen}.
For existing assets, $\mathcal{X}$-Part supports generative shape decomposition guided by bounding-box prompts~\cite{yan2025xpart}, while P$^3$-SAM enables automatic and point-prompted 3D part segmentation~\cite{ma2025p3sam}.

\begin{figure*}[h]
    \centering
\includegraphics[width=0.8\textwidth]{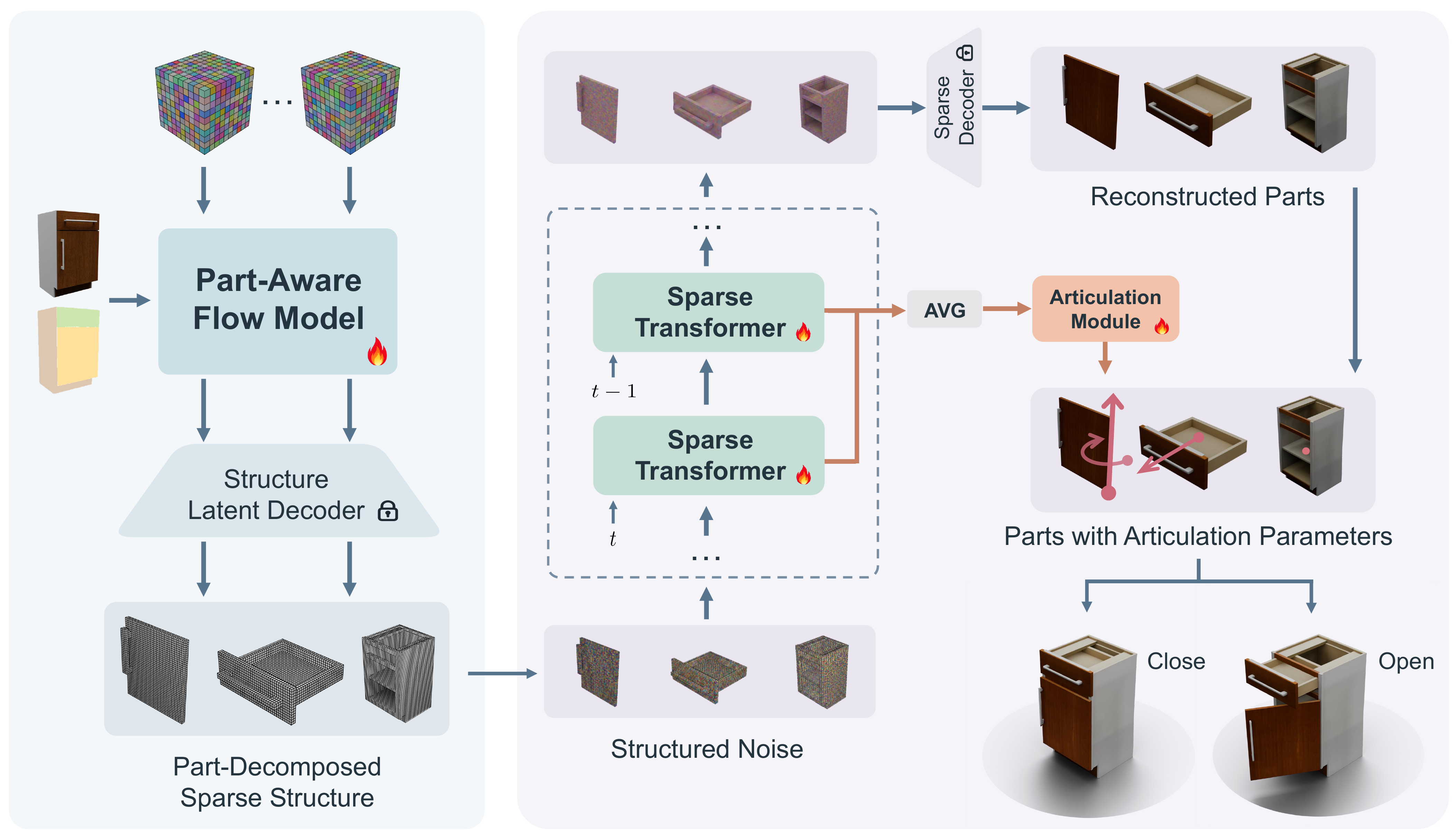}
    \caption{
    Pipeline of \methodname. Stage~1 predicts a part-decomposed sparse structure from a single image using a Part-Aware Flow Model. Stage~2 refines it into detailed 3D part representations via sparse-transformer denoising, while an articulation module aggregates multi-step features to estimate joint parameters for each part. The predicted joint parameters together with the reconstructed part geometries form the final articulated object.
    }
    \label{fig:pipe}
\end{figure*}

\section{Preliminary: TRELLIS — A 3D-Native Generative Model}
Pretrained object-level 3D generative models provide strong priors for synthesizing high-quality geometry and appearance from limited observations. Given the scarcity of large-scale datasets for articulated objects, leveraging such pretrained 3D priors forms an effective foundation for our approach. In particular, \methodname\ builds upon TRELLIS~\cite{xiang2024trellis}, a two-stage 3D-native generative framework for producing detailed 3D representations of static objects from a single image or a text prompt.

\paragraph{Stage 1: Sparse Structure Generation.}
TRELLIS adopts a two-stage pipeline that generates 3D objects from coarse structure to fine-grained details. In the first stage, object geometry is represented as a sparse binary occupancy grid at a resolution of $64^3$. A variational autoencoder (VAE) encodes this sparse structure into a compact spatial latent representation:
\begin{equation}
\small
z = \text{E}_{\text{ss}}(x) \in \mathbb{R}^{16 \times 16 \times 16 \times c}, \quad
\hat{x} = \text{D}_{\text{ss}}(z),
\end{equation}
where $x \in \mathbb{R}^{64 \times 64 \times 64}$ denotes the input occupancy grid, $\text{E}_{\text{ss}}$ and $\text{D}_{\text{ss}}$ are the encoder and decoder, and $z$ is the resulting latent tensor. A rectified flow model $\mathcal{RF}_{\text{ss}}$ is trained in this latent space to enable conditional generation of coarse object structures.

\paragraph{Stage 2: Geometry and Appearance Generation.}
To model fine-grained geometry and appearance, TRELLIS constructs a structured latent feature volume (SLAT):
\begin{equation}
\small
\mathcal{F} = \{ (\mathbf{x}_i, \mathbf{f}_i) \}_{i=1}^L,
\quad
\mathbf{x}_i \in \{0,\dots,63\}^3,
\quad
\mathbf{f}_i \in \mathbb{R}^d,
\end{equation}
where each occupied voxel location $\mathbf{x}_i$ is associated with a feature vector $\mathbf{f}_i$ extracted from image-based features. This sparse representation encodes detailed appearance information only at occupied locations.

The sparse feature volume is further encoded into a latent representation using a sparse encoder--decoder architecture:
\begin{equation}
\small
\mathcal{Z} = \text{E}_{\text{SLAT}}(\mathcal{F})
= \{ (\mathbf{x}_i, \mathbf{z}_i) \}_{i=1}^L,
\quad
\hat{\mathcal{F}} = \text{D}_{\text{SLAT}}(\mathcal{Z}),
\end{equation}
where $\mathbf{z}_i \in \mathbb{R}^d$ denotes the latent feature at location $\mathbf{x}_i$. A second rectified flow model, $\mathcal{RF}_{\text{SLAT}}$, is trained in this sparse latent space to enable conditional generation of detailed geometry and appearance, after which the decoder reconstructs the final 3D representation.

\section{Method}

We propose a two-stage generative pipeline for reconstructing articulated objects from a single-view input (Fig.~\ref{fig:pipe}). 
Stage~1 predicts a part-decomposed sparse structure using part-level latents and mask-based part conditioning. 
Stage~2 synthesizes detailed part geometry and appearance conditioned on this structure and predicts articulation parameters from intermediate denoising features.

\subsection{Problem Definition}
\label{sec:problem_def}
Given a single RGB image $\mathcal{I}$ as input, our goal is to reconstruct a complete articulated object representation $\mathcal{O}$ that is (i) geometrically consistent, (ii) visually faithful to the input image, and (iii) physically plausible with respect to its articulation structure. We assume that objects admit a tree-structured articulation graph with a single root, which covers many articulated objects in practice.

Formally, we represent an articulated object as $\mathcal{O}=\{\mathcal{P}_i\}_{i=1}^{K}$, with each part $\mathcal{P}_i=(\mathcal{G}_i,\mathcal{A}_i)$, where $K$ denotes the number of semantic parts including the fixed root, $\mathcal{G}_i$ denotes geometry and appearance, and $\mathcal{A}_i$ encodes articulation information.

We consider three joint types between each part and its parent: \emph{revolute}, \emph{prismatic}, and \emph{fixed}. Following prior work~\cite{liu2024cage,liu2024singapo}, we parameterize the articulation of part $i$ as
\begin{equation}
\small
\mathcal{A}_i = \{\tau_i,\
\mathbf{s}_i,\
\mathbf{o}_i,\
\mathbf{u}_i,\
\boldsymbol{\rho}_i,\
\mathbf{q}_i\},
\end{equation}
where $\tau_i$ is the joint type, $\mathbf{s}_i$ is the semantic label, $\mathbf{o}_i\in\mathbb{R}^{3}$ is the joint origin in the world coordinate system, $\mathbf{u}_i\in\mathbb{R}^{3}$ is the joint axis, $\boldsymbol{\rho}_i=[\rho_i^{\min},\rho_i^{\max}]\in\mathbb{R}^{2}$ specifies the motion range, and $\mathbf{q}_i$ indicates the parent of part $i$ in the articulation tree. For numerical stability, we normalize the predicted axis direction as $\mathbf{u}_i \leftarrow \mathbf{u}_i / \|\mathbf{u}_i\|_2$.

\subsection{Stage~1: Part-Decomposed Structure Generation}
\label{sec:stage1}

\begin{figure}[h]
    \centering
\includegraphics[width=0.65\linewidth]{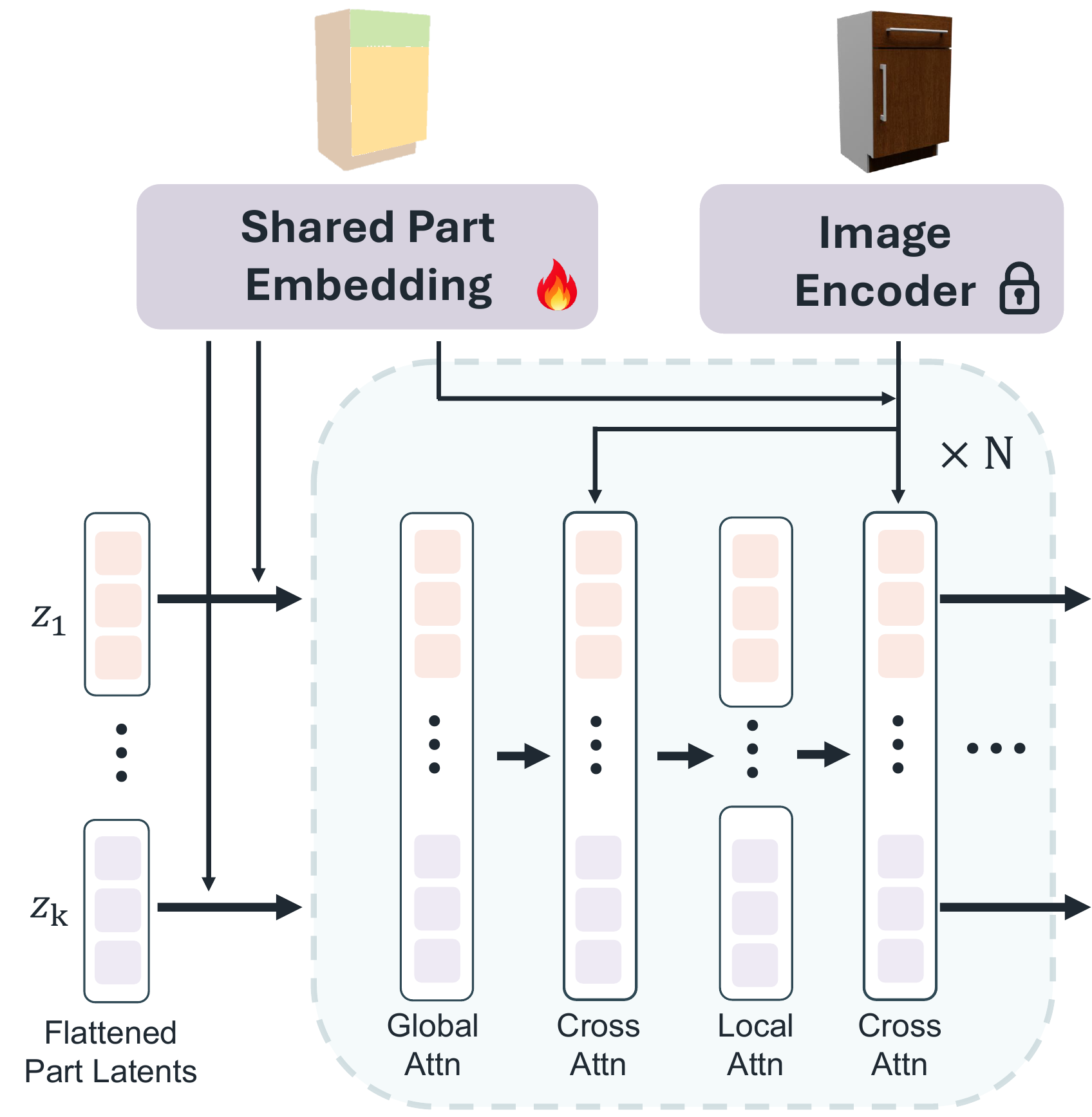}
    \caption{Architecture of the Stage~1 model. To accommodate multi-part inputs, a subset of attention layers in the original TRELLIS network is repurposed to perform within-part local attention. The shared part embedding is added to each  part latent and corresponding image tokens to inject part identity. With these modifications, the model is fine-tuned on the target dataset.
    }
    \label{fig:stage1}
\end{figure}

We first generate a part-decomposed sparse structure for the object. Specifically, we represent the object by a set of part-level latent tokens
$\mathcal{Z}_{\mathrm{global}}=\{\mathbf{z}_{i}\}_{i=1}^{K}$, where $\mathbf{z}_{i}$ encodes the sparse structure of part $i$, and $K$ is the number of parts. To distinguish different parts, we introduce learnable part-identity embeddings $\mathbf{E}\in\mathbb{R}^{K_{\max}\times d_p}$, where $K_{\max}$ is the maximum number of parts. For each token of part $i$, we add the corresponding identity embedding to its latent feature:
$\mathbf{z}'_i = \mathbf{z}_i + \mathbf{E}[i]$. The architecture is shown in Fig.~\ref{fig:stage1}.

\paragraph{Part-aware Denoising Transformer.} 
TRELLIS is pretrained to model a single coherent object. To adapt it to part-decomposed generation, we repurpose a subset of attention layers to perform within-part attention, while keeping the remaining layers global to model inter-part interactions. This allows the model to reuse pretrained object-level priors at the part level while still reasoning over the full object structure. Implementation details are provided in Sec.~\ref{sec:imple_details}.

\paragraph{Within-part Local Attention.}
In the repurposed layers, self-attention is restricted to tokens from the same part. For part $i$, the within-part attention is
\begin{equation}
\small
\mathbf{H}_i = \mathbf{z}'_i + \mathrm{Attn}\!\left(
\mathbf{z}'_i \mathbf{W}^{p}_{Q},\,
\mathbf{z}'_i \mathbf{W}^{p}_{K},\,
\mathbf{z}'_i \mathbf{W}^{p}_{V}
\right),
\end{equation}
where $\mathrm{Attn}(\mathbf{Q}, \mathbf{K}, \mathbf{V})=\mathrm{Softmax}\!\left(\frac{\mathbf{Q}\mathbf{K}^{\top}}{\sqrt{d}}\right)\mathbf{V}$.
The output is then passed through an FFN with a residual connection:
$\mathbf{H}_i \leftarrow \mathbf{H}_i + \mathrm{FFN}(\mathbf{H}_i).
$

\paragraph{Mask-based Part Conditioning.}
To provide explicit spatial guidance for part decomposition, we condition the model on a 2D part mask $\mathcal{M}$. The mask assigns a discrete part index to each image location. We embed these indices using the same Stage~1 part-identity embedding $\mathbf{E}$, ensuring consistent part semantics between image-space conditioning and part-level latent tokens. Given $\mathcal{M}\in\{0,1,\dots,K_{\max}-1\}^{h\times w}$, we construct a dense part-embedding map
\begin{equation}
\small
\mathbf{P}_{u,v} = \mathbf{E}\!\left[\mathcal{M}_{u,v}\right] \in \mathbb{R}^{d_p}.
\end{equation}
This map is downsampled and added to DINOv2 image features~\cite{oquab2024dinov},
which are injected into the denoising transformer through cross-attention.
The downsampled mask therefore acts as a coarse localization cue
rather than a pixel-accurate geometric constraint.
Since mask indices have no intrinsic semantic order, we randomly permute them during training
to improve robustness and controllability.

\subsection{Stage~2: Part Structure Conditioned Generation and Articulation Prediction}
\label{sec:stage2}

\subsubsection{Structure-Conditioned Object Generation}

Given the part-level decomposition from Stage~1, we obtain part-wise \emph{sparse voxels} that encode coarse geometry and segmentation but lack fine details and appearance. Stage~2 synthesizes detailed \textbf{geometry} and \textbf{appearance} for each part conditioned on these sparse structures.

Our design is related to OmniPart~\cite{yang2025omnipart}, which conditions part synthesis on AABBs predicted by a separate module. In contrast, we directly use the Stage~1 sparse voxels as part-wise structural conditions, avoiding an additional box predictor. Following TRELLIS and OmniPart, we flatten sparse structured latents into a 1D token sequence with positional encodings. Since the Stage~1 sparse voxels are part-aware, tokens from all parts are concatenated into a \emph{global object token sequence} for object-level denoising (Fig.~\ref{fig:stage2}, top). A learnable \emph{part embedding}, separate from the Stage~1 part identity embedding, is added to each token to preserve part identity while enabling global object reasoning. The frozen TRELLIS sparse-latent decoder $D_{\text{SLAT}}$ then reconstructs per-part 3D representations, yielding part geometries $\{\mathcal{G}_i\}_{i=1}^{K}$.

\subsubsection{Articulation Regression}
\label{sec:arti_regression}
\paragraph{Articulation Parameterization.}
Many datasets annotate fine-grained sub-components, such as handles or knobs, as separate nodes, although they are often connected to parent parts by \emph{fixed} joints and introduce no additional degrees of freedom. Unlike prior methods~\cite{liu2024cage,liu2024singapo,wu2025dipo} that keep these nodes in the articulation graph, we collapse fixed-joint nodes into their parents and merge their geometry and appearance into the corresponding semantic parts. This preserves the kinematic degrees of freedom while removing redundant nodes and edges.
\begin{figure}[h]
    \centering
\includegraphics[width=0.90\linewidth]{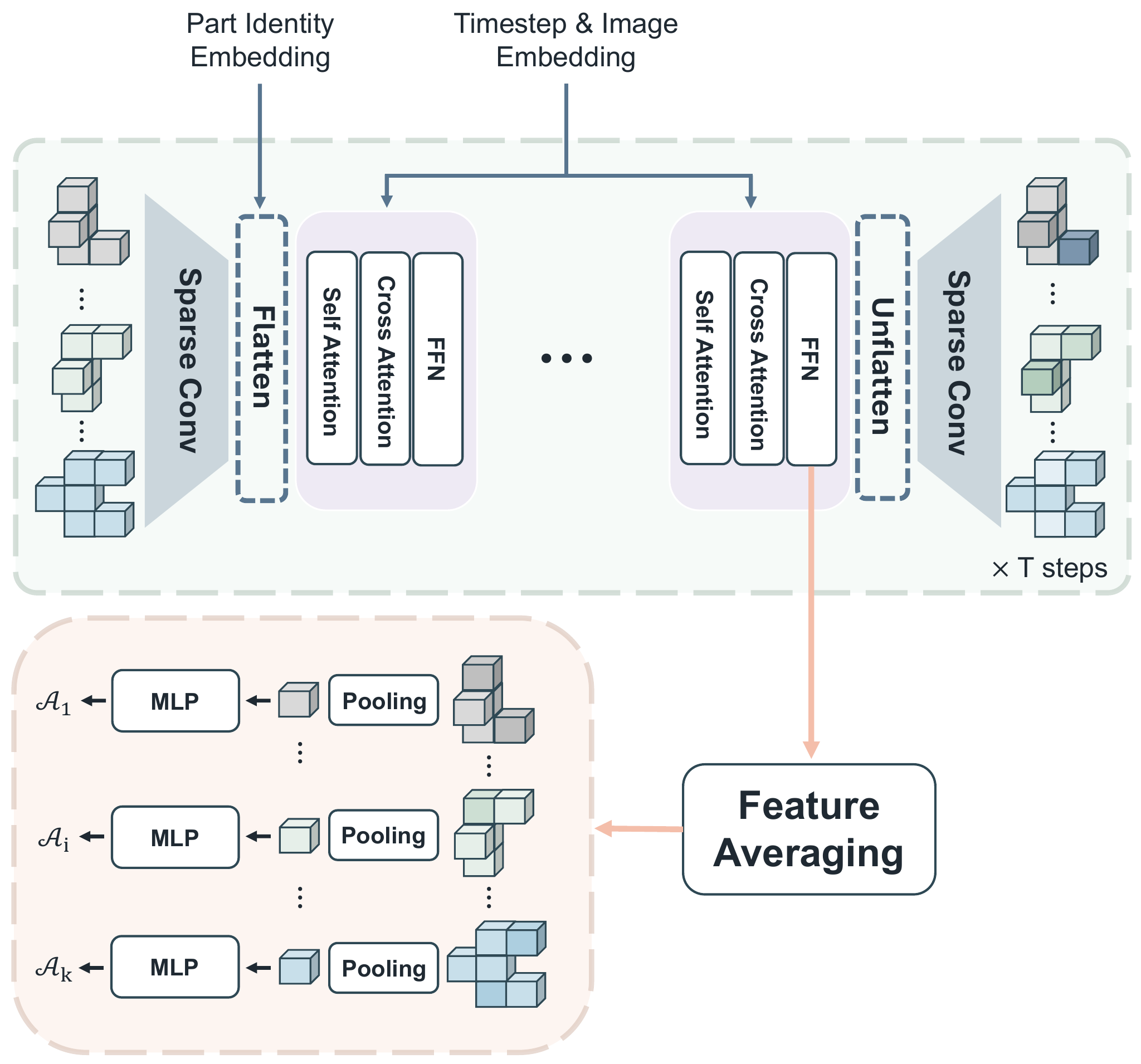}
    \caption{Illustration of the Stage~2 pipeline. We cache a subset of intermediate features during part-token denoising. For each part, these features are averaged over selected inference timesteps and then pooled over the token dimension. A learned MLP then predicts the articulation parameters.
    }
    \label{fig:stage2}
\end{figure}
Following the experimental setting of prior works~\cite{liu2024singapo,wu2025dipo}, we assume a depth-1 articulation tree, where all movable parts are directly attached to a single fixed root. 
This setting covers common articulated objects such as cabinets and bottles, and
many more commonly encountered in embodied environments.
We therefore assign each part a semantic label, e.g., \texttt{base}, \texttt{door}, or \texttt{drawer}, treat \texttt{base} as the root, and define all non-root articulation parameters relative to it. 
Once the \texttt{base} part is identified, the parent index $\mathbf{q}_i$ is determined for all non-root parts. 
A preliminary extension to depth-$N$ structures is explored and discussed in Appendix~\ref{app:complex_graph}.

\paragraph{Predicting Joint Parameters from Part Tokens.}
The denoising backbone in our second stage operates on part-centric structured latents. After flattening, each part $i$ is represented by a token sequence $\mathcal{Z}_i \in \mathbb{R}^{L_i \times D}$, where $L_i$ may vary across parts and $D$ is the token dimension. Let $\mathbf{H}_{i}^{t}\in\mathbb{R}^{L_{i}\times D}$ denote the corresponding token features from the last transformer block at diffusion step $t$.

\paragraph{Multi-step feature aggregation.}
Inspired by Diffusion Hyperfeatures~\cite{luo2023dhf}, we aggregate denoiser features
from multiple timesteps, which provide complementary cues (Fig.~\ref{fig:stage2}, bottom).
Earlier denoising features emphasize coarse object and part layout,
whereas later features encode finer part geometry.
Because the backbone jointly denoises the concatenated tokens of all parts
while part embeddings preserve part identity,
these features also combine part-specific shape cues with cross-part relations,
both of which are relevant to joint estimation.
For selected steps $\mathcal{T}$, we average part-level token features over time:
\begin{equation}
\small
\overline{\mathbf{H}}_{i} = \frac{1}{|\mathcal{T}|}\sum_{t\in\mathcal{T}}\mathbf{H}_{i}^{t}
\in \mathbb{R}^{L_i \times D}.
\end{equation}
We obtain a compact part descriptor by concatenating token-wise mean and max pooling,
\begin{equation}
\small
\overline{\mathbf{h}}_{i} =
\Big[
\mathrm{MeanPool}(\overline{\mathbf{H}}_{i})
\ \big\|\
\mathrm{MaxPool}(\overline{\mathbf{H}}_{i})
\Big]
\in \mathbb{R}^{2D},
\end{equation}
and predict articulation parameters with an MLP:
\begin{equation}
\small
\hat{\mathcal{A}}_i =
\{\hat{\tau}_i,\hat{\mathbf{s}}_i,\hat{\mathbf{o}}_i,\hat{\mathbf{u}}_i,\hat{\boldsymbol{\rho}}_i\}
= g_{\phi}(\overline{\mathbf{h}}_i).
\end{equation}
In practice, we use the last $S$ denoising steps, $\mathcal{T}=\{T-S+1,\ldots,T\}$, which reuses existing denoising features with negligible overhead.

\subsection{Training}
\label{sec:training}
\methodname{} is trained at the object level: each sample contains all parts of an articulated object, represented by concatenated part latents.
In Stage~1, the flow matching loss is applied over these latents to learn the part-decomposed sparse structure. 
In Stage~2, we jointly optimize flow matching and articulation regression:
\begin{equation}
\mathcal{L} = \mathcal{L}_{\text{fm}} + \lambda \, \mathcal{L}_{\text{art}}.
\end{equation}
Articulation parameters are predicted from intermediate part-token activations (Sec.~\ref{sec:arti_regression}) and supervised by an $\ell_2$ loss, where categorical attributes are represented as real-valued vectors following SINGAPO~\cite{liu2024singapo}:
\begin{equation}
\small
\mathcal{L}_{\text{art}} = \sum_{i=1}^{K} \left\| \hat{\mathcal{A}}_i - \mathcal{A}_i \right\|_2^2,
\end{equation}

where $\mathcal{A}_i$ and $\hat{\mathcal{A}}_i$ are the ground-truth and predicted articulation parameter vectors for part $i$. Since flow-matching training samples one timestep per iteration, $\mathcal{L}_{\text{art}}$ is computed from the corresponding single-step features rather than the multi-step aggregated features used at inference.

\section{Experiments}

\subsection{Implementation Details and Evaluation Setup.}
\textit{Implementation Details.}
\label{sec:imple_details}
Our Part-Aware Flow Model is initialized from the Stage~1 checkpoint of TRELLIS,
and the sparse transformer is initialized from the Stage~2 checkpoint of OmniPart~\cite{yang2025omnipart}.
We train the two stages separately, using a learning rate of $1\times 10^{-4}$ for 40K steps
on 4 NVIDIA A800 GPUs (80GB), with an effective batch size of 32.
The articulation head $g_{\phi}$ is a six-layer MLP with hidden width 512.
At inference time, we use a CFG scale of 7 for both stages and sample 25 denoising steps per stage.
In the Part-Aware Flow Model, global and within-part local attention are interleaved by depth:
even-indexed layers perform global attention, while odd-indexed layers perform within-part attention.
For articulation prediction, we aggregate features from the last 20 denoising steps.
Following the SINGAPO protocol, we use 20 rendered views per training object
and two randomly sampled, distinct views per test object;
the evaluation viewpoints differ from the training views.
For part masks, we consider two sources: ground-truth annotations when available,
and automatically generated masks from a VLM-guided prompting pipeline
followed by SAM2~\cite{ravi2024sam2} refinement
(VLM+SAM; see Appendix~\ref{app:vlm_mask}).

\textit{Dataset and Metrics.}
We follow the setup of SINGAPO~\cite{liu2024singapo} and DIPO~\cite{wu2025dipo}: models are trained on PartNet-Mobility~\cite{xiang2020sapien} and evaluated on both PartNet-Mobility and ACD~\cite{iliash2024s2o}, using the standard train/test splits adopted by SINGAPO. For articulated object generation, we report four metrics used in prior work: \textbf{dgIoU}$\downarrow$, the generalized IoU between predicted and ground-truth part bounding boxes; \textbf{dcDist}$\downarrow$, the Euclidean distance between predicted and ground-truth part centers; \textbf{dCD}$\downarrow$, the Chamfer distance between predicted and ground-truth meshes; and \textbf{AOR}$\downarrow$, the Average Overlapping Ratio for inter-part collision/validity. We compute these metrics in both resting and articulated states, denoted by the prefixes \textbf{RS-} and \textbf{AS-}, respectively. Beyond geometric fidelity and articulation accuracy, we evaluate \emph{visual} consistency with the input image, \textit{i.e.}, whether the generated object matches the observed instance. Following FreeArt3D~\cite{chen2025freeart3d}, we report an image-based \textbf{CLIP} similarity score~\cite{radford2021learning}. Specifically, we render the generated and ground-truth textured meshes from five random viewpoints under six articulation states, and compute the CLIP similarity between each rendering and the corresponding GT image.

\textit{Baselines.}
We compare with representative articulated object generation methods:
SINGAPO~\cite{liu2024singapo}, DIPO~\cite{wu2025dipo},
Articulate-Anything~\cite{le2024articulate}, PhysX-3D~\cite{cao2025physx},
PhysX-Anything~\cite{physxanything}, ArtFormer~\cite{Su_2025_CVPR},
and FreeArt3D~\cite{chen2025freeart3d}.
SINGAPO, DIPO, and Articulate-Anything are retrieval-based pipelines:
SINGAPO and DIPO infer a part connectivity graph and abstract part attributes
before retrieving part meshes from a database, while Articulate-Anything uses a VLM-driven pipeline
to retrieve, assemble, and articulate assets from a library.
DIPO additionally uses images of two articulation states,
providing motion cues unavailable to our single-image method.
PhysX-3D and PhysX-Anything target physics-grounded, simulation-ready asset generation
by producing explicit geometry together with kinematic and physical attributes.
ArtFormer represents an articulated object as a tokenized kinematic tree
and uses a transformer to generate geometry codes and kinematic relations
for controllable articulated object synthesis.
Finally, FreeArt3D is a training-free, per-instance optimization method
that jointly optimizes geometry and articulation from multiple observations
of the same object under different articulation states.
Unlike the single-image methods, DIPO and FreeArt3D use additional observations. We do not include direct comparisons with ArtiLatent or UniArt due to
unavailable codes and checkpoints.

\vspace{-4mm}
\subsection{General Results}
\begin{figure*}[p]
    \centering
\includegraphics[width=1.0\textwidth]{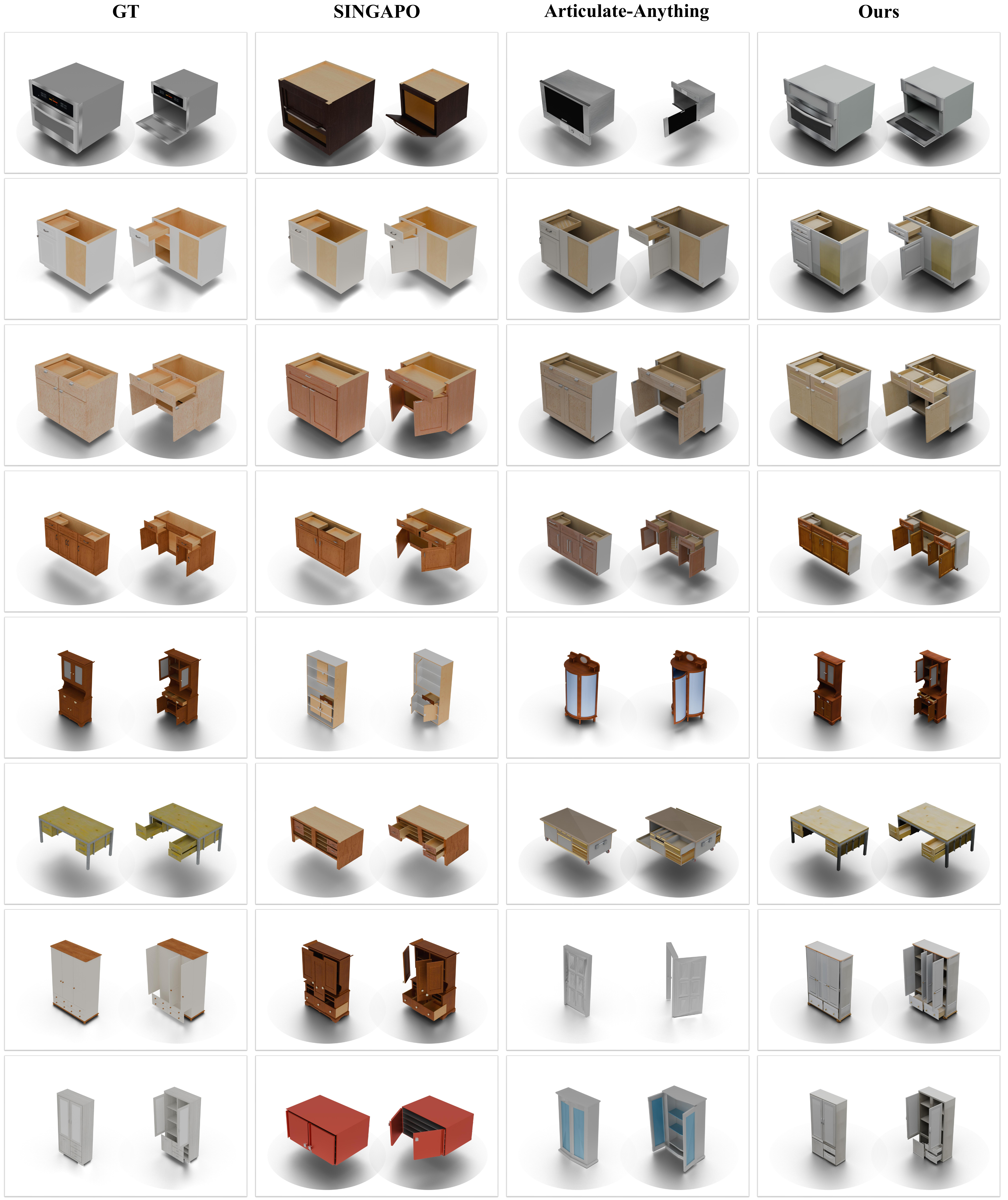}
\vspace{-20pt}
    \caption{
    Qualitative comparison on the PartNet-Mobility~\cite{xiang2020sapien}
    and ACD~\cite{iliash2024s2o} datasets, including ground truth (GT), SINGAPO~\cite{liu2024singapo},
    Articulate-Anything~\cite{le2024articulate}, and \methodname.
    }
    \label{fig:cmp_baseline}
\end{figure*}

We present qualitative comparisons on PartNet-Mobility and ACD in Fig.~\ref{fig:cmp_baseline}, and quantitative results in Tab.~\ref{tab:metrics_pm} and Tab.~\ref{tab:metrics_acd} (Appendix), respectively. A separate qualitative comparison with DIPO is provided in Appendix~\ref{app:dipo_qualitative}. For qualitative evaluation, all methods are visualized as textured meshes using the same Blender-based rendering pipeline.

\begin{table*}[ht]
\centering
\small
\setlength{\tabcolsep}{6pt}
\renewcommand{\arraystretch}{1.15}
\caption{Comparison with baselines on the PartNet-Mobility Dataset~\cite{xiang2020sapien}.
The GT-mask result is an oracle evaluation of the part-aware generation module;
VLM+SAM is the practical automatic-mask setting.
}
\begin{tabular}{l|cccccccc}
\toprule
\textbf{Methods$\backslash$Metric} & 
RS-$d_{\text{gIoU}}\downarrow$ & AS-$d_{\text{gIoU}}\downarrow$ & RS-$d_{\text{cDist}}\downarrow$ & AS-$d_{\text{cDist}}\downarrow$ & RS-$d_{\text{CD}}\downarrow$&
AS-$d_{\text{CD}}\downarrow$ & AOR$\downarrow$ & CLIP$\uparrow$ \\
\midrule
SINGAPO~\cite{liu2024singapo}& 0.5119 & 0.5166 & 0.1177 & 0.1505 & 0.0040 & 0.0202 & 0.0097 & 0.8779 \\
DIPO~\cite{wu2025dipo}& 0.6221 & 0.6310 & 0.0564 & 0.1119 & 0.0035 & 0.0188 & 0.0573 & 0.8774 \\
ArtFormer~\cite{Su_2025_CVPR}  & 1.3165 & 1.3212 & 0.3542 & 0.5099 & 0.0504 & 0.1849 & 0.0191 & 0.7877 \\
PhysX-3D~\cite{cao2025physx} & 0.7610 & 0.7637 & 0.3233 & 0.4562 & 0.0115 & 0.1204 & 0.2125 & 0.7972 \\
PhysX-Anything~\cite{physxanything} & 0.9596 & 0.9661 & 0.2538 & 0.4034 & 0.0097 & 0.1106 & 0.0056 & 0.8859 \\
Arti-Anything~\cite{le2024articulate} & 0.6865 & 0.6969 & 0.1386 & 0.3414 & 0.0107 & 0.1230 & 0.0056 & 0.8795 \\
\midrule
Ours (VLM+SAM Mask) & 0.2485 & 0.2736 & 0.0707 & 0.1191 & 0.0019 & 0.0176 & 0.0233  & 0.8996 \\
\midrule
\textbf{Ours} (GT Mask) &
\textbf{0.1395} & \textbf{0.1695} & \textbf{0.0202} & \textbf{0.0653} &
\textbf{0.0016} & \textbf{0.0139} & \textbf{0.0011} & \textbf{0.9053} \\
\bottomrule
\end{tabular}
\label{tab:metrics_pm}
\end{table*}

\begin{table*}[ht]
\centering
\small
\setlength{\tabcolsep}{6pt}
\renewcommand{\arraystretch}{1.15}
\caption{Ablation study with different configurations on the PartNet-Mobility dataset~\cite{xiang2020sapien}.}
\begin{tabular}{l|cccccccc}
\toprule
\textbf{Methods$\backslash$Metric} & 
RS-$d_{\text{gIoU}}\downarrow$ & AS-$d_{\text{gIoU}}\downarrow$ & RS-$d_{\text{cDist}}\downarrow$ & AS-$d_{\text{cDist}}\downarrow$ & RS-$d_{\text{CD}}\downarrow$&
AS-$d_{\text{CD}}\downarrow$ & AOR$\downarrow$ & CLIP$\uparrow$ \\
\midrule
$S$ = 1  & 0.1734 & 0.1999 & 0.0300 & 0.0966 & 0.0017 & 0.0220 & 0.0088  & 0.8903 \\
$S$ = 15  & 0.1737 & 0.2002 & 0.0320 & 0.0762 & 0.0018 & 0.0145 & 0.0090 & 0.8911 \\
$S$ = 20  & 0.1395 & 0.1695 & 0.0202 & 0.0653 & 0.0016 & 0.0139 & 0.0011  & 0.9053 \\
Stage 1-Articulation  & 0.1615 & 0.1878 & 0.0302 & 0.0889 & 0.0016 & 0.0210 & 0.0068  & 0.8903 \\
FM Articulation & 0.3615 & 0.4132 & 0.1263 & 0.3019 & 0.0021 & 0.1100 & 0.0190 & 0.8803 \\
\midrule
\textbf{Ours} (GT Mask)  & \textbf{0.1395} & \textbf{0.1695} & \textbf{0.0202} & \textbf{0.0653} & \textbf{0.0016} & \textbf{0.0139} & \textbf{0.0011} & \textbf{0.9053} \\
\bottomrule
\end{tabular}
\label{tab:ablation}
\end{table*}

\textit{Qualitative Results.}
As shown in Fig.~\ref{fig:cmp_baseline}, our method produces detailed geometry
and preserves major visual attributes of the input image in many cases.
Compared with the displayed baselines, our reconstructions often exhibit finer geometric details, more accurate part shapes, and plausible articulation structures.
Retrieval-based methods such as SINGAPO and Articulate-Anything often struggle to recover instance-specific geometry and appearance. SINGAPO performs part retrieval without explicitly conditioning on image appearance, leading to noticeable deviations from the target instance (\eg, the first row). Articulate-Anything relies heavily on vision-language reasoning, which can introduce semantic misclassification; for example, in the seventh row, a white cabinet is incorrectly identified as a ``door,'' causing downstream errors in structure and articulation prediction.

\begin{figure*}[p]
    \centering
\includegraphics[width=0.95\textwidth]{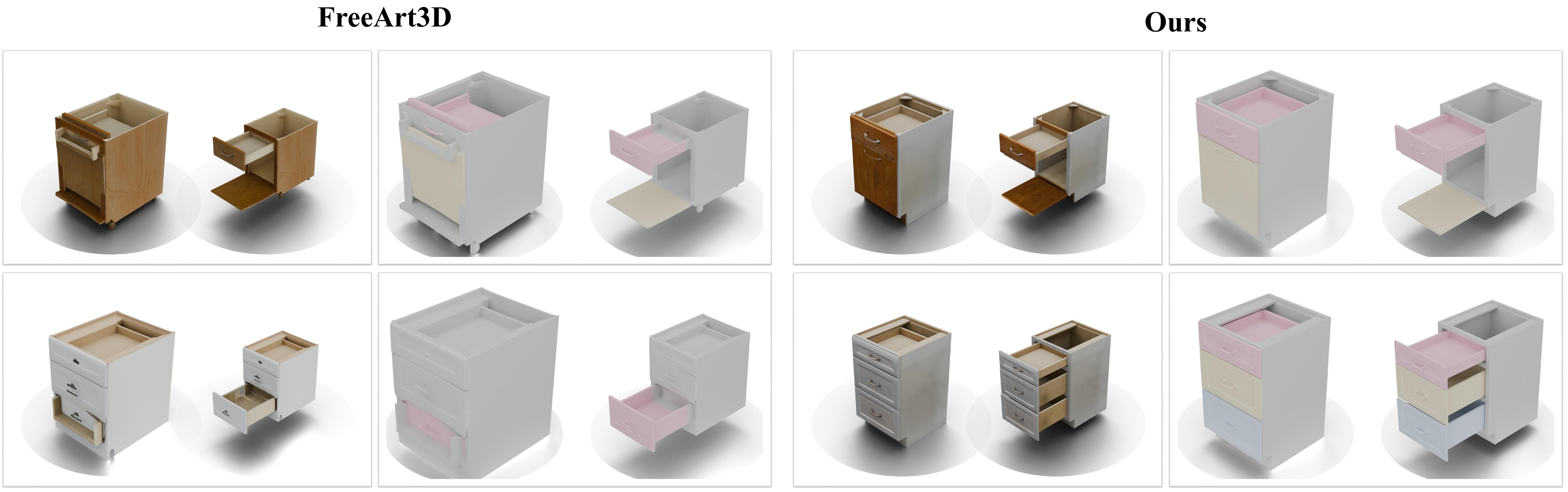}
\vspace{-10pt}
\caption{Qualitative comparison with FreeArt3D~\cite{chen2025freeart3d}.
We visualize the geometry of different parts using distinct colors.}
    \label{fig:cmp_fa3d}
\end{figure*}

\textit{Comparison with Optimization-based Methods.}
We further compare with FreeArt3D, which reconstructs articulated objects from multiple images captured under different articulation states via SDS-based optimization. In Fig.~\ref{fig:cmp_fa3d}, we visualize both the final textured meshes and the raw part geometries, with each part rendered in a distinct color. Due to space limitations, the corresponding ground-truth shapes are provided in the Appendix (Fig.~\ref{fig:cmp_fa3d_gt}). FreeArt3D often produces noisy or distorted part geometries and inconsistent part boundaries, leading to fragmented or inaccurate decompositions. In contrast, our method yields cleaner part geometries and more coherent part segmentation, despite using only a single input image. This highlights the advantage of our generative formulation over per-instance optimization, even when the latter has access to multiple articulated views.

\textit{Quantitative Results.}
The GT-mask oracle achieves the strongest overall performance
across the evaluated motion-space metrics, while the practical VLM+SAM setting retains clear geometry gains but exhibits a noticeable gap on several structure and articulation metrics.
By directly synthesizing 3D geometry rather than retrieving parts from a database, our model produces more instance-specific shapes, leading to improved resting-state accuracy
(\eg, lower RS-$d_{\text{CD}}$).
The articulated-state results evaluate the combined effect of generated geometry and predicted motion;
direct joint-axis estimation is discussed separately in Appendix~\ref{app:joint_param_eval}.
Additional analysis of sensitivity to mask quality is provided in
Appendix~\ref{app:additional_ablations}.
We report the runtime of all methods in Tab.~\ref{tab:runtime}.

\begin{table}[ht]
\centering
\footnotesize
\setlength{\tabcolsep}{3.5pt}
\renewcommand{\arraystretch}{1.15}
\caption{Runtime comparison with baselines on a single NVIDIA A800 GPU.} %
\resizebox{\linewidth}{!}{
\begin{tabular}{ccccccc}
\toprule
Ours & Arti-Anything & ArtFormer & SINGAPO & PhysX-3D &PhysX-Anything &FreeArt3D \\
\midrule
\textasciitilde15s & \textasciitilde3min & \textasciitilde155s & \textasciitilde8s & \textasciitilde96s 
&  \textasciitilde 20min
&  \textasciitilde 10min \\
\bottomrule
\end{tabular}
}
\label{tab:runtime}
\end{table}

\vspace{-4mm}
\subsection{Ablation Studies}

In \methodname, articulation parameters are predicted by a lightweight head that decodes denoiser features from the Sparse Transformer in Stage~2. The head aggregates features from multiple denoising steps and regresses articulation parameters from the fused representation. To evaluate this design, we conduct ablations on PartNet-Mobility along three axes: (1) the number of denoising steps used for feature aggregation; (2) predicting articulation from Stage~2 features versus Stage~1 features from the Part-Aware Flow Model; and (3) replacing the deterministic regression head with a generative, flow-matching-based articulation predictor.

\textit{Number of Denoising Steps.}
As shown in Tab.~\ref{tab:ablation}, aggregating features from multiple denoising steps consistently improves articulation prediction. We evaluate $S=\{1,15,20\}$ denoising steps. Using a single step provides limited temporal context and leads to less stable predictions, while aggregating more steps improves performance with diminishing returns. Overall, using \textbf{20} denoising steps achieves the best performance and is adopted as our default setting.

\textit{Stage~1 vs.\ Stage~2 Features.}
Predicting articulation from Stage~1 features (``Stage 1-Articulation'') performs substantially worse than using Stage~2 features. This reflects the different roles of the two stages: Stage~1 features mainly encode coarse part structure, whereas Stage~2 features are refined through object-level denoising and contain richer geometric and semantic information, making them better suited for articulation estimation.

\textit{Regression vs.\ Generative Articulation prediction.}
Replacing the deterministic MLP regression head with a flow-matching-based generative predictor (``FM Articulation'') does not yield consistent improvements on PartNet-Mobility, while introducing additional training and inference complexity. We therefore adopt the simpler regression-based articulation module in all experiments.

\begin{figure*}[p]
    \centering
\includegraphics[width=0.85\textwidth]{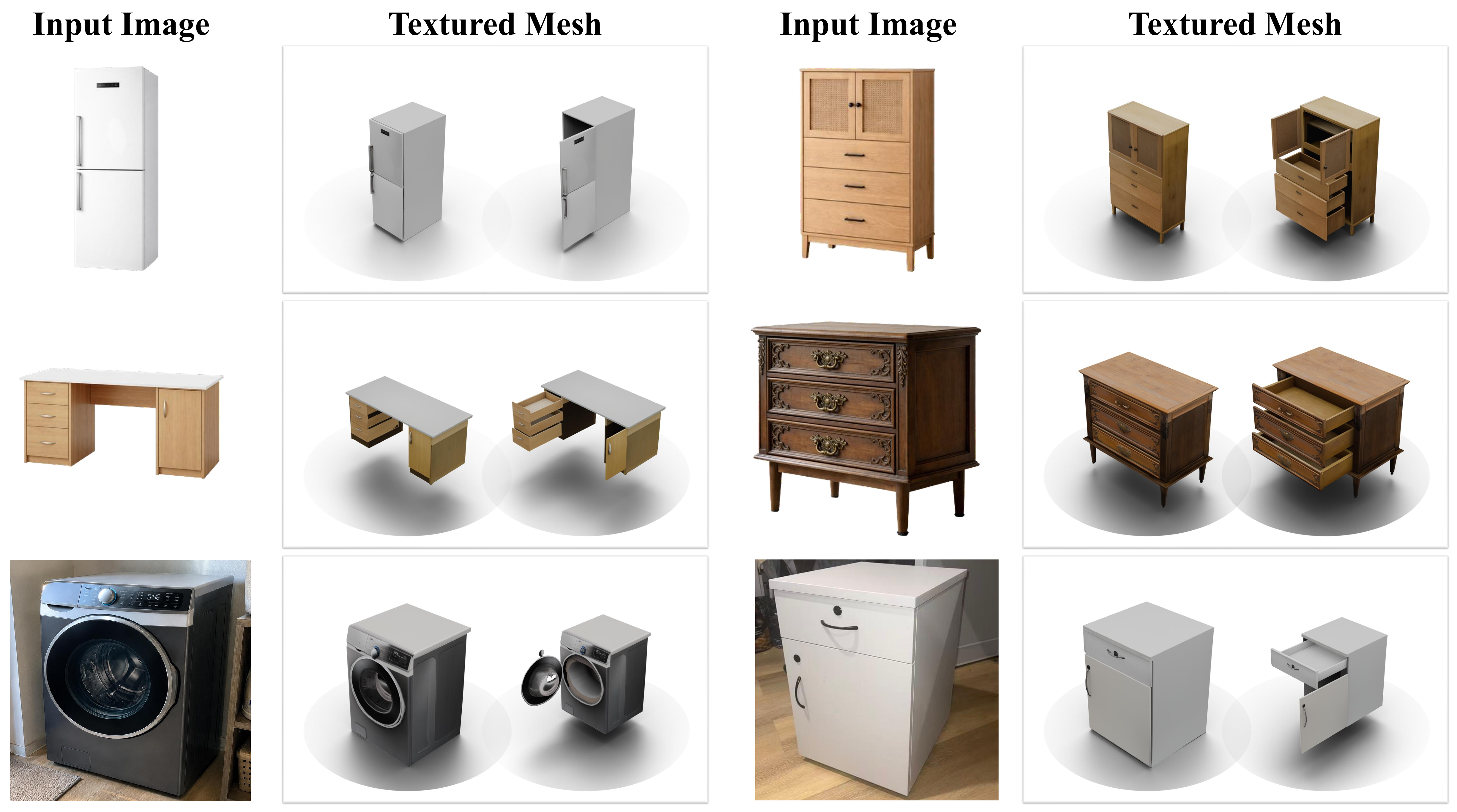}
\vspace{-10pt}
    \caption{Qualitative results on the in-the-wild images collected from the internet or produced by ChatGPT. 
    }
    \label{fig:real_world}
\end{figure*}

\begin{figure*}[p]
    \centering
\includegraphics[width=0.85\textwidth]{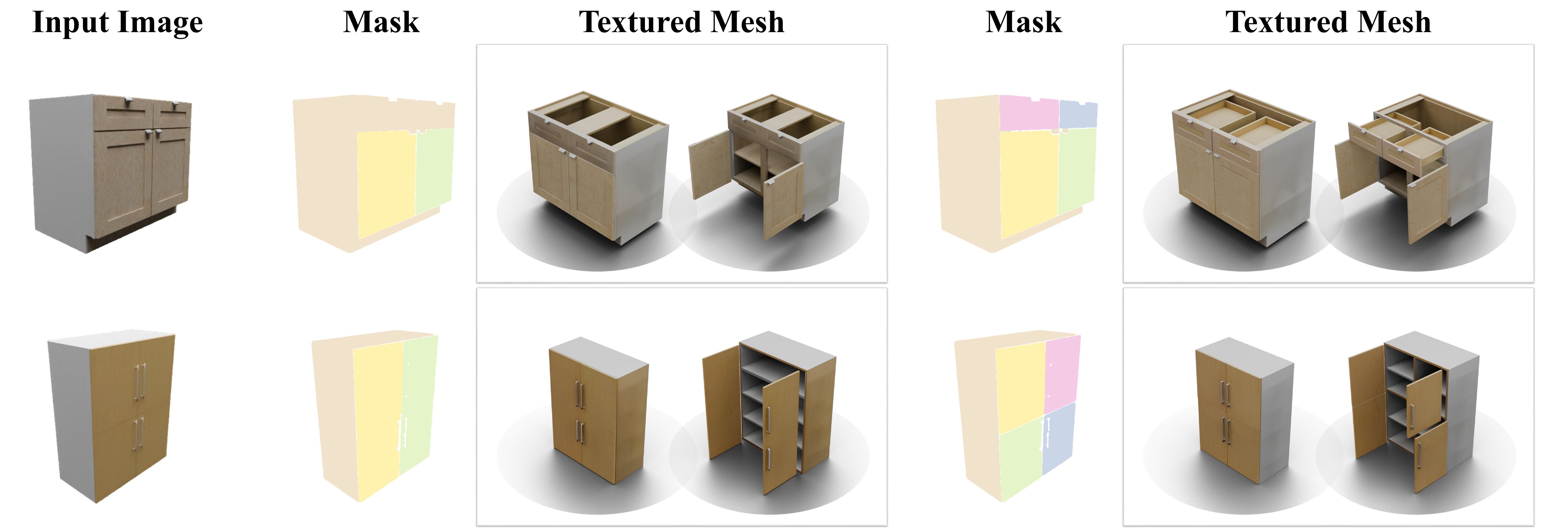}
\vspace{-10pt}
    \caption{Qualitative results using the same images but with different masks as control signals.
    }
    \label{fig:mask_control}
\end{figure*}

\begin{figure}[!t]
    \centering
    \begin{minipage}[t]{0.50\textwidth}
        \centering
        \includegraphics[width=0.90\linewidth]{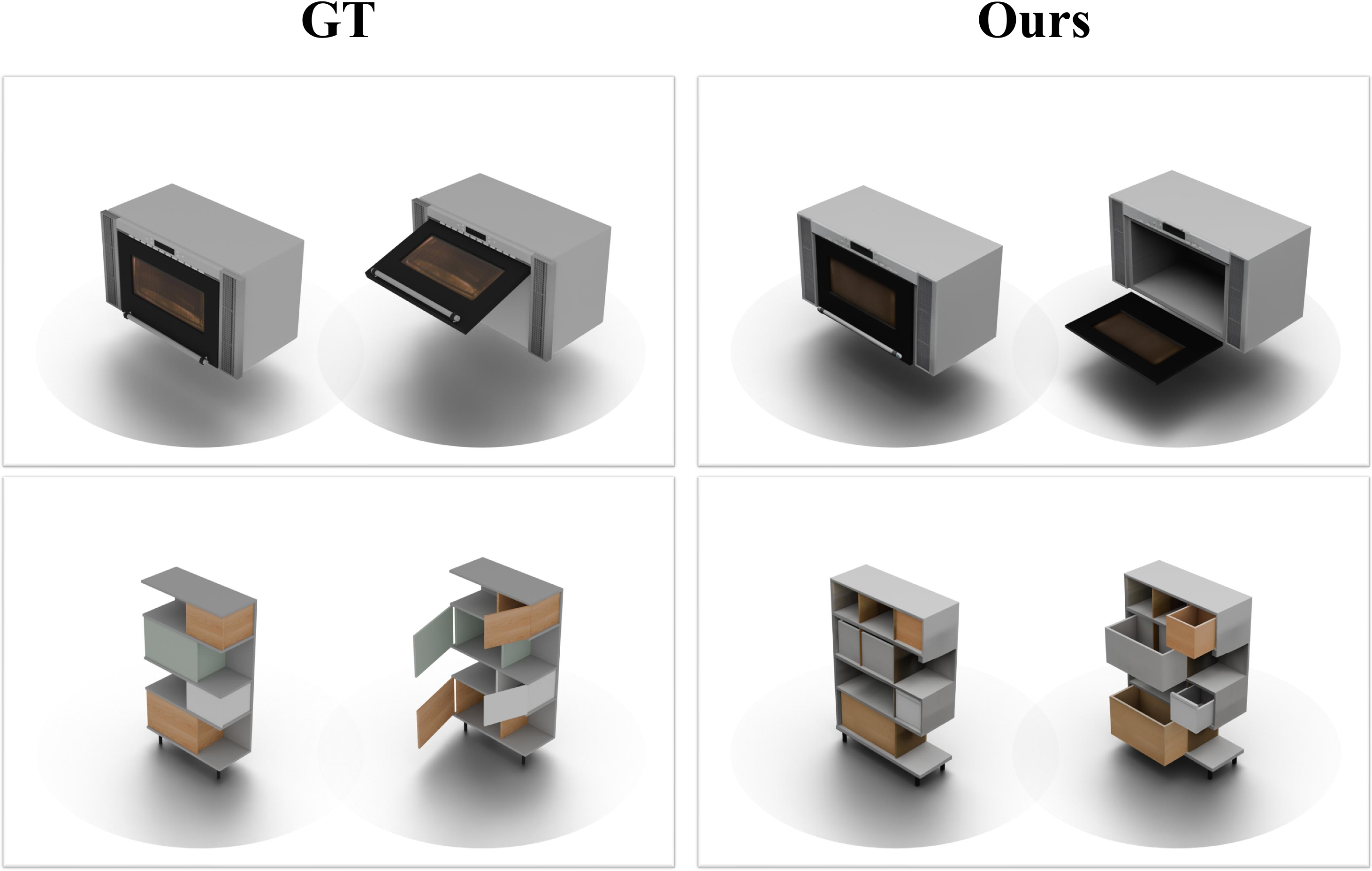}
        \captionof{figure}{Representative joint-estimation failures. Each row compares
        ground-truth (GT) articulation states with the corresponding generated result.}
        
        \label{fig:failure_joint_estimation}
    \end{minipage}
\end{figure}

\subsection{Real-World Articulated Object Generation}
Beyond standard benchmarks, we qualitatively evaluate \methodname on a small, curated set of in-the-wild-style inputs, including ChatGPT-synthesized images and real photos of common articulated objects such as cabinets, desks, refrigerators, and washing machines, under the same single-image setting. As shown in Fig.~\ref{fig:real_world}, the method can produce coherent part decompositions, plausible geometry, and visually consistent textures for these examples. Additional in-the-wild examples and a simulator demonstration are provided in Appendix~\ref{app:additional_applications}.
Although corresponding ground-truth 3D assets are unavailable, these curated examples demonstrate the practical applicability of \methodname{} to in-the-wild inputs.

\subsection{Mask-Controlled Generation}
A single input image may admit multiple valid part decompositions, leading to different yet plausible articulation structures. To enable controllable generation under such ambiguity, we use a semantic part mask as an explicit conditioning signal. By specifying the target semantic parts, the mask guides both the decomposition and the resulting articulated object. Fig.~\ref{fig:mask_control} illustrates this effect using two alternative masks for each image. For example, in the first row, different masks guide the model to generate either (i) two doors only or (ii) two doors with two drawers, while preserving consistent appearance and plausible joint configurations. This mask-based conditioning enables controllable generation and reduces ambiguity in part structure and articulation for real-world images.

\begin{figure}[!t]
\centering
\begin{minipage}[t]{0.50\textwidth}
    \centering
    \includegraphics[width=0.90\linewidth]{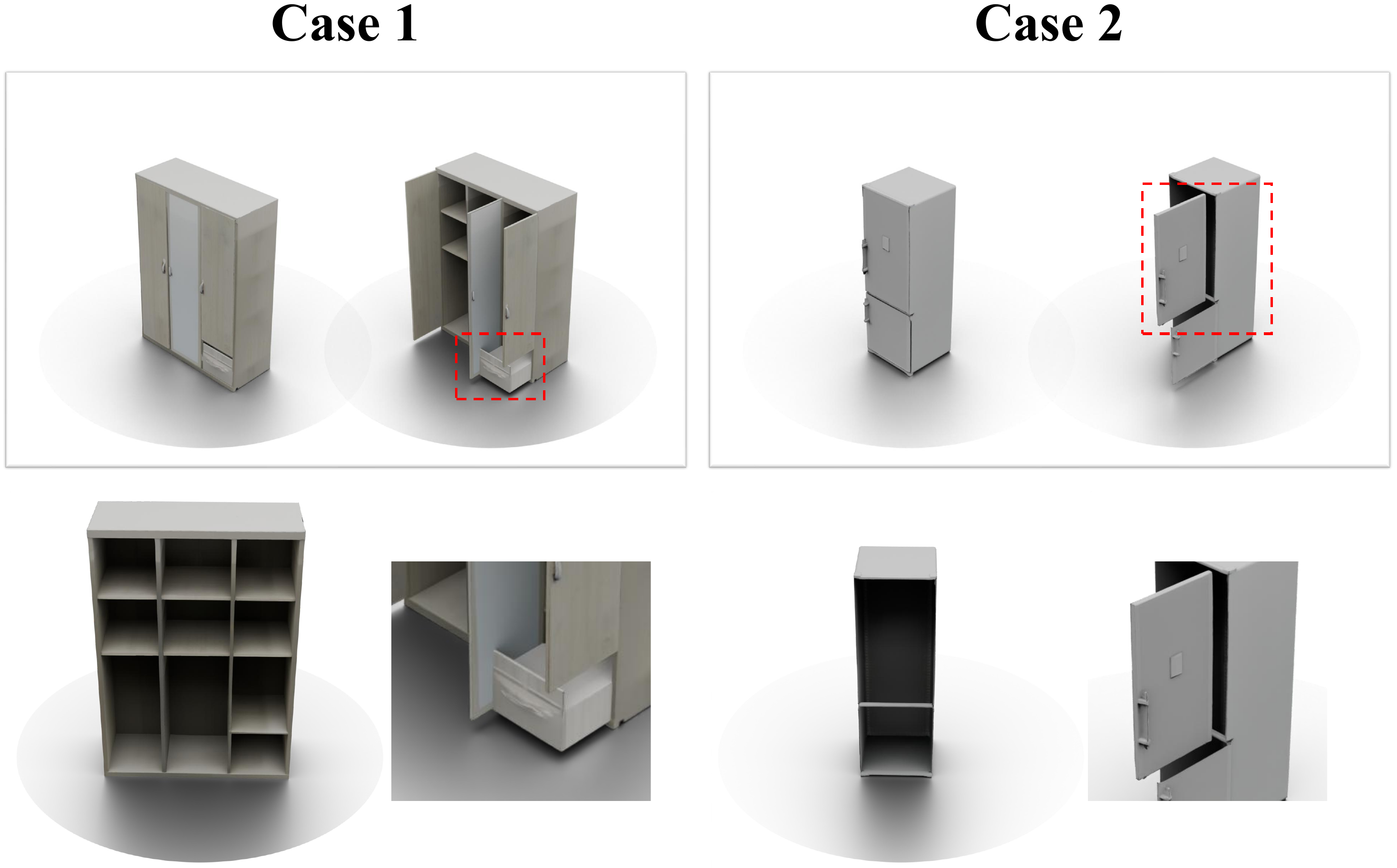}
    \captionof{figure}{Representative physical-validity and appearance failures.
    Case~1 shows a collision between the door and drawer; Case~2 shows a door
    detached from the \texttt{base}. Red boxes highlight the affected regions.
    The bottom-left \texttt{base}-only renderings expose black interior regions
    caused by texture misalignment.}
    \label{fig:failure_physical_appearance}
\end{minipage}
\end{figure}

\section{Representative Failure Cases}
\label{sec:failure_cases}

Figure~\ref{fig:failure_joint_estimation} compares predicted articulation states
with ground-truth references. Because single-view generation is inherently ambiguous,
the ground truth represents one valid articulation rather than the only plausible solution.
Thus, differences from the ground truth may reflect plausible alternatives, while larger
discrepancies can reveal errors in geometry generation or joint estimation.

Figure~\ref{fig:failure_physical_appearance} highlights two limitations of our method. 
The physical failures reflect the absence of explicit physical constraints and could potentially be mitigated by simulator-based validity feedback during generation or refinement. 
The appearance artifacts in the \texttt{base}-only renderings arise from texture misalignment in occluded interiors, as discussed in Sec.~\ref{sec:limitations}.

\vspace{-2.5mm}
\section{Limitations and Conclusion}
\label{sec:limitations}

\textit{Category, part-count, and input-state scope.}
Training and quantitative evaluation concentrate on seven household furniture and appliance categories. The method struggles with objects containing many parts because such examples are rare in existing articulated-object datasets, whose scale is roughly 3K objects in our setting after augmentation. Following SINGAPO, we also assume resting-state or mostly closed inputs; performance may degrade for arbitrary articulation states, whose handling would require jointly estimating the observed articulation state and canonical joint parameters.

\textit{Unseen or occluded parts.}
We assume that relevant functional parts are visible in the input image. When components are occluded or only become visible under certain articulation states, such as a drawer behind a closed cabinet door, the model may fail to infer and generate them. For such unseen regions, the model relies on learned geometric and structural priors rather than direct visual evidence; consequently, the generated components may be plausible but are not guaranteed to match the specific instance. Incorporating occlusion reasoning and stronger priors for completing unseen components is a promising direction.

\textit{Kinematic scope.}
We adopt a depth-1 kinematic formulation suited to common categories
with shallow articulation trees. A preliminary graph-prediction extension
in Appendix~\ref{app:complex_graph} shows limited generalization under the current limited-data scale. Extending to richer kinematic structures, such as coupled
joints or closed-chain mechanisms remains an important direction for future work.

\textit{Texture misalignment.}
\methodname{} may exhibit texture misalignment or degradation in some cases (Fig.~\ref{fig:failure_physical_appearance}), mainly due to the TRELLIS texture reconstruction pipeline, where textures are back-projected from rendered Gaussian splatting results and can contain errors. 
The dark appearance of hollow or occluded interiors introduces ambiguity into appearance supervision. 
This limitation is orthogonal to our articulation modeling and could be alleviated by native texture-generation backbones, such as TRELLIS2~\cite{xiang2025trellis2}.

\textit{Conclusion.} 
We present a part-centric framework for single-view articulated object generation that decomposes objects into movable parts and reconstructs their geometry, appearance, and articulation. Experiments demonstrate improved articulation accuracy, high-quality part geometry, and strong visual consistency over competitive baselines.

\begin{acks}
This work was supported by the Guangdong S\&T Programme under Grant 2024B0101040004, the Shenzhen Science and Technology Program under Grant KJZD20240903104008012, the Shenzhen Science and Technology Program under Grant ZDCY20250901113000001, and the CUHK--CUHK(SZ)--GDSTC Joint Collaboration Fund under Grant No. 2025A0505000053.
\end{acks}

\clearpage
\bibliographystyle{ACM-Reference-Format}
\bibliography{sample-base}

@String{CVPR = "Proceedings of the IEEE/CVF Conference on Computer Vision and Pattern Recognition" }

@String{Computer = "{IEEE} Computer" }

@String{Springer = "Springer-Verlag" }

@article{tang2024partpacker,
  title={Efficient Part-level 3D Object Generation via Dual Volume Packing},
  author={Tang, Jiaxiang and Lu, Ruijie and Li, Zhaoshuo and Hao, Zekun and Li, Xuan and Wei, Fangyin and Song, Shuran and Zeng, Gang and Liu, Ming-Yu and Lin, Tsung-Yi},
  journal={arXiv preprint arXiv:2506.09980},
  year={2025}
}

@article{lin2025partcrafter,
  title={PartCrafter: Structured 3D Mesh Generation via Compositional Latent Diffusion Transformers}, 
  author={Yuchen Lin and Chenguo Lin and Panwang Pan and Honglei Yan and Yiqiang Feng and Yadong Mu and Katerina Fragkiadaki},
  year={2025},
  journal={arXiv preprint arXiv:2506.05573},
}

@article{yang2025omnipart,
        title={Omnipart: Part-aware 3d generation with semantic decoupling and structural cohesion},
        author={Yang, Yunhan and Zhou, Yufan and Guo, Yuan-Chen and Zou, Zi-Xin and Huang, Yukun and Liu, Ying-Tian and Xu, Hao and Liang, Ding and Cao, Yan-Pei and Liu, Xihui},
        journal={arXiv preprint arXiv:2507.06165},
        year={2025}
}

@article{cao2025physx,
  title={PhysX-3D: Physical-Grounded 3D Asset Generation},
  author={Cao, Ziang and Chen, Zhaoxi and Pan, Liang and Liu, Ziwei},
  journal={arXiv preprint arXiv:2507.12465},
  year={2025}
}

@article{physxanything,
  title={PhysX-Anything: Simulation-Ready Physical 3D Assets from Single Image},
  author={Cao, Ziang and Hong, Fangzhou and Chen, Zhaoxi and Pan, Liang and Liu, Ziwei},
  journal={arXiv preprint arXiv:2511.13648},
  year={2025}
}

@article{lu2025dreamart,
  title={Dreamart: Generating interactable articulated objects from a single image},
  author={Lu, Ruijie and Liu, Yu and Tang, Jiaxiang and Ni, Junfeng and Wang, Yuxiang and Wan, Diwen and Zeng, Gang and Chen, Yixin and Huang, Siyuan},
  journal={arXiv preprint arXiv:2507.05763},
  year={2025}
}

@article{chen2024urdformer,
  title={{URDFormer}: A Pipeline for Constructing Articulated Simulation Environments from Real-World Images},
  author={Chen, Zoey and Walsman, Aaron and Memmel, Marius and Mo, Kaichun and Fang, Alex and Vemuri, Karthikeya and Wu, Alan and Fox, Dieter and Gupta, Abhishek},
  journal={arXiv preprint arXiv:2405.11656},
  year={2024}
}

@article{jin2025uniart,
  title={UniArt: Unified 3D Representation for Generating 3D Articulated Objects with Open-Set Articulation},
  author={Jin, Bu and Li, Weize and Gu, Songen and Zheng, Yupeng and Zheng, Yuhang and Zhou, Zhengyi and Yao, Yao},
  journal={arXiv preprint arXiv:2511.21887},
  year={2025}
}

@inproceedings{chen2025artilatent,
  title={ArtiLatent: Realistic Articulated 3D Object Generation via Structured Latents},
  author={Chen, Honghua and Lan, Yushi and Chen, Yongwei and Pan, Xingang},
  booktitle={Proceedings of the SIGGRAPH Asia 2025 Conference Papers},
  year={2025},
  doi={10.1145/3757377.3763922}
}

@article{wu2025dipo,
title={DIPO: Dual-State Images Controlled Articulated Object Generation Powered by Diverse Data},
author={Wu, Ruqi and Wang, Xinjie and Liu, Liu and Guo, Chunle and Qiu, Jiaxiong and Li, Chongyi and Huang, Lichao and Su, Zhizhong and Cheng, Ming-Ming},
journal={Advances in Neural Information Processing Systems 39},
year={2025},
}

@article{gao2024meshart,
  title={MeshArt: Generating Articulated Meshes with Structure-guided Transformers},
  author={Gao, Daoyi and Siddiqui, Yawar and Li, Lei and Dai, Angela},
  journal={arXiv preprint arXiv:2412.11596},
  year={2024}
}

@InProceedings{Su_2025_CVPR,
    author    = {Su, Jiayi and Feng, Youhe and Li, Zheng and Song, Jinhua and He, Yangfan and Ren, Botao and Xu, Botian},
    title     = {ArtFormer: Controllable Generation of Diverse 3D Articulated Objects},
    booktitle = {Proceedings of the IEEE/CVF Conference on Computer Vision and Pattern Recognition (CVPR)},
    month     = {June},
    year      = {2025},
    pages     = {1894-1904}
}

@article{he2025spark,
    title={SPARK: Sim-ready Part-level Articulated Reconstruction with VLM Knowledge}, 
    author={Yumeng He and Ying Jiang and Jiayin Lu and Yin Yang and Chenfanfu Jiang},
    journal={arXiv preprint arXiv:2512.01629},
    year={2025}
}

@article{liu2024singapo,
  title={{SINGAPO}: Single Image Controlled Generation of Articulated Parts in Object},
  author={Liu, Jiayi and Iliash, Denys and Chang, Angel X and Savva, Manolis and Mahdavi-Amiri, Ali},
  journal={arXiv preprint arXiv:2410.16499},
  year={2024}
}

@InProceedings{chen2025freeart3d,
  title = {FreeArt3D: Training-Free Articulated Object Generation using 3D Diffusion},
  author = {Chen, Chuhao and Liu, Isabella and Wei, Xinyue and Su, Hao and Liu, Minghua},
  booktitle = {SIGGRAPH Asia 2025 Conference Papers},
  year = {2025}
}

@article{li2025art,
  title     = {ART: Articulated Reconstruction Transformer},
  author    = {Li, Zizhang and Zhang, Cheng and Li, Zhengqin and Howard-Jenkins, Henry and Lv, Zhaoyang and Geng, Chen and Wu, Jiajun and Newcombe, Richard and Engel, Jakob and Dong, Zhao},
  journal   = {arXiv preprint arXiv:2512.14671},
  year      = {2025}
}

@article{liu2024cage,
  title={CAGE: Controllable Articulation GEneration},
  author={Liu, Jiayi and Tam, Hou In Ivan and Mahdavi-Amiri, Ali and Savva, Manolis},
  journal={CVPR},
  year={2024}
}

@article{wang2025kinematif,
      title={Kinematify: Open-Vocabulary Synthesis of High-DoF Articulated Objects}, 
      author={Jiawei Wang and Dingyou Wang and Jiaming Hu and Qixuan Zhang and Jingyi Yu and Lan Xu},
      year={2025},
      journal={arXiv preprint arXiv:2511.01294},
}

@article{mandi2024real2code,
  title={Real2Code: Reconstruct Articulated Objects via Code Generation},
  author={Mandi, Zhao and Weng, Yijia and Bauer, Dominik and Song, Shuran},
  journal={arXiv preprint arXiv:2406.08474},
  year={2024}
}

@article{yuan2025larmlargearticulatedobjectreconstruction,
      title={LARM: A Large Articulated-Object Reconstruction Model}, 
      author={Yuan, Sylvia and Shi, Ruoxi and Wei, Xinyue and Zhang, Xiaoshuai and Su, Hao and Liu, Minghua},
      journal={arXiv preprint arXiv:2511.11563},
      year={2025},
}

@article{liu2025videoartgs,
  title={VideoArtGS: Building Digital Twins of Articulated Objects from Monocular Video},
  author={Liu, Yu and Jia, Baoxiong and Lu, Ruijie and Gan, Chuyue and Chen, Huayu and Ni, Junfeng and Zhu, Song-Chun and Huang, Siyuan},
  journal={arXiv preprint arXiv:2509.17647},
  year={2025}
}

@inproceedings{liu2025building,
  title={Building Interactable Replicas of Complex Articulated Objects via Gaussian Splatting},
  author={Liu, Yu and Jia, Baoxiong and Lu, Ruijie and Ni, Junfeng and Zhu, Song-Chun and Huang, Siyuan},
  booktitle={The Thirteenth International Conference on Learning Representations},
  year={2025},
}

@inproceedings{jiayi2023paris,
  title={PARIS: Part-level Reconstruction and Motion Analysis for Articulated Objects},
  author={Liu, Jiayi and Mahdavi-Amiri, Ali and Savva, Manolis},
  booktitle={ICCV},
  year={2023}
}

@article{mao2022multiscan,
  title={MultiScan: Scalable RGBD scanning for 3D environments with articulated objects},
  author={Mao, Yongsen and Zhang, Yiming and Jiang, Hanxiao and Chang, Angel X and Savva, Manolis},
  journal={NeurIPS},
  year={2022}
}

@inproceedings{hsu2023ditto,
  title={Ditto in the house: Building articulation models of indoor scenes through interactive perception},
  author={Hsu, Cheng-Chun and Jiang, Zhenyu and Zhu, Yuke},
  booktitle={ICRA},
  year={2023},
}

@article{nie2022structure,
  title={Structure from action: Learning interactions for articulated object 3d structure discovery},
  author={Nie, Neil and Gadre, Samir Yitzhak and Ehsani, Kiana and Song, Shuran},
  journal={arXiv preprint arXiv:2207.08997},
  year={2022}
}

@inproceedings{song2024reacto,
  title={REACTO: Reconstructing Articulated Objects from a Single Video},
  author={Song, Chaoyue and Wei, Jiacheng and Foo, Chuan Sheng and Lin, Guosheng and Liu, Fayao},
  booktitle=CVPR,
  year={2024}
}

@article{lin2025splart,
  title={SplArt: Articulation Estimation and Part-Level Reconstruction with 3D Gaussian Splatting},
  author={Lin, Shengjie and Fang, Jiading and Irshad, Muhammad Zubair and Guizilini, Vitor Campagnolo and Ambrus, Rares Andrei and Shakhnarovich, Greg and Walter, Matthew R},
  journal={arXiv preprint arXiv:2506.03594},
  year={2025}
}

@inproceedings{jiang2022ditto,
  title={Ditto: Building digital twins of articulated objects from interaction},
  author={Jiang, Zhenyu and Hsu, Cheng-Chun and Zhu, Yuke},
  booktitle=CVPR,
  year={2022}
}

@article{lei2023nap,
  title={Nap: Neural 3d articulated object prior},
  author={Lei, Jiahui and Deng, Congyue and Shen, William B and Guibas, Leonidas J and Daniilidis, Kostas},
  journal={Advances in Neural Information Processing Systems},
  year={2023}
}

@article{le2024articulate,
    author = {Le, Long and Xie, Jason and Liang, William and Wang, Hung-Ju and Yang, Yue and Ma, Yecheng Jason and Vedder, Kyle and Krishna, Arjun and Jayaraman, Dinesh and Eaton, Eric},
    title = {Articulate-anything: Automatic modeling of articulated objects via a vision-language foundation model},
    journal={ICLR},
    year = {2025}
}

@article{qiu2025articulate,
  title={Articulate AnyMesh: Open-Vocabulary 3D Articulated Objects Modeling},
  author={Qiu, Xiaowen and Yang, Jincheng and Wang, Yian and Chen, Zhehuan and Wang, Yufei and Wang, Tsun-Hsuan and Xian, Zhou and Gan, Chuang},
  journal={arXiv preprint arXiv:2502.02590},
  year={2025}
}

@article{lian2025infinite,
  title={Infinite Mobility: Scalable High-Fidelity Synthesis of Articulated Objects via Procedural Generation},
  author={Lian, Xinyu and Yu, Zichao and Liang, Ruiming and Wang, Yitong and Luo, Li Ray and Chen, Kaixu and Zhou, Yuanzhen and Tang, Qihong and Xu, Xudong and Lyu, Zhaoyang and others},
  journal={arXiv preprint arXiv:2503.13424},
  year={2025}
}

@article{zhou2026articraft,
  title={Articraft: An Agentic System for Scalable Articulated 3D Asset Generation},
  author={Zhou, Matt and Li, Ruining and Lyu, Xiaoyang and Song, Zhaomou and Huang, Zhening and Zheng, Chuanxia and Rupprecht, Christian and Vedaldi, Andrea and Wu, Shangzhe},
  journal={arXiv preprint arXiv:2605.15187},
  year={2026}
}

@Article{kerbl3Dgaussians,
      author       = {Kerbl, Bernhard and Kopanas, Georgios and Leimk{\"u}hler, Thomas and Drettakis, George},
      title        = {3D Gaussian Splatting for Real-Time Radiance Field Rendering},
      journal      = {ACM Transactions on Graphics},
      number       = {4},
      volume       = {42},
      month        = {July},
      year         = {2023}
}

@article{xiang2020sapien,
  title={Sapien: A simulated part-based interactive environment},
  author={Xiang, Fanbo and Qin, Yuzhe and Mo, Kaichun and Xia, Yikuan and Zhu, Hao and Liu, Fangchen and Liu, Minghua and Jiang, Hanxiao and Yuan, Yifu and Wang, He and others},
  journal={CVPR},
  year={2020}
}

@article{xiang2024trellis,
    title   = {Structured 3D Latents for Scalable and Versatile 3D Generation},
    author  = {Xiang, Jianfeng and Lv, Zelong and Xu, Sicheng and Deng, Yu and Wang, Ruicheng and Zhang, Bowen and Chen, Dong and Tong, Xin and Yang, Jiaolong},
    journal = {arXiv preprint arXiv:2412.01506},
    year    = {2024}
}

@article{
    xiang2025trellis2,
    title={Native and Compact Structured Latents for 3D Generation},
    author={Xiang, Jianfeng and Chen, Xiaoxue and Xu, Sicheng and Wang, Ruicheng and Lv, Zelong and Deng, Yu and Zhu, Hongyuan and Dong, Yue and Zhao, Hao and Yuan, Nicholas Jing and Yang, Jiaolong},
    journal = {arXiv preprint arXiv:2512.14692},
    year={2025}
}

@article{li2025triposg,
  title={TripoSG: High-Fidelity 3D Shape Synthesis using Large-Scale Rectified Flow Models},
  author={Li, Yangguang and Zou, Zi-Xin and Liu, Zexiang and Wang, Dehu and Liang, Yuan and Yu, Zhipeng and Liu, Xingchao and Guo, Yuan-Chen and Liang, Ding and Ouyang, Wanli and others},
  journal={arXiv preprint arXiv:2502.06608},
  year={2025}
}

@article{ma2025p3sam,
  title={P3-sam: Native 3d part segmentation},
  author={Ma, Changfeng and Li, Yang and Yan, Xinhao and Xu, Jiachen and Yang, Yunhan and Wang, Chunshi and Zhao, Zibo and Guo, Yanwen and Chen, Zhuo and Guo, Chunchao},
  journal={arXiv preprint arXiv:2509.06784},
  year={2025}
}

@article{yan2025xpart,
  title={X-Part: high fidelity and structure coherent shape decomposition},
  author={Yan, Xinhao and Xu, Jiachen and Li, Yang and Ma, Changfeng and Yang, Yunhan and Wang, Chunshi and Zhao, Zibo and Lai, Zeqiang and Zhao, Yunfei and Chen, Zhuo and others},
  journal={arXiv preprint arXiv:2509.08643},
  year={2025}
}

@article{chen2024partgen,
  title={PartGen: Part-level 3D Generation and Reconstruction with Multi-View Diffusion Models},
  author={Minghao Chen and Roman Shapovalov and Iro Laina and Tom Monnier and Jianyuan Wang and David Novotny and Andrea Vedaldi},
  journal={arXiv preprint arXiv:2412.18608},
  year={2024}
}

@article{chen2025autopartgen,
  title={AutoPartGen: Autogressive 3D Part Generation and Discovery},
  author={Minghao Chen and Jianyuan Wang and Roman Shapovalov and Tom Monnier and Hyunyoung Jung and Dilin Wang and Rakesh Ranjan and Iro Laina and Andrea Vedaldi},
  journal={arXiv preprint arXiv:2507.13346},
  year={2025}
}

@article{lai2025hunyuan3d25highfidelity3d,
      title={Hunyuan3D 2.5: Towards High-Fidelity 3D Assets Generation with Ultimate Details}, 
      author={Tencent Hunyuan3D Team},
      year={2025},
      journal={arXiv preprint arXiv:2506.16504},
      archivePrefix={arXiv},
      primaryClass={cs.CV},
}

@article{hunyuan3d22025tencent,
    title={Hunyuan3D 2.0: Scaling Diffusion Models for High Resolution Textured 3D Assets Generation},
    author={Tencent Hunyuan3D Team},
    year={2025},
    journal={arXiv preprint arXiv:2501.12202},
    archivePrefix={arXiv},
    primaryClass={cs.CV}
}

@article{yang2024hunyuan3d,
    title={Hunyuan3D 1.0: A Unified Framework for Text-to-3D and Image-to-3D Generation},
    author={Tencent Hunyuan3D Team},
    year={2024},
    journal={arXiv preprint arXiv:2411.02293},
    archivePrefix={arXiv},
    primaryClass={cs.CV}
}

@article{seed3D1.0,
  title={Seed3D 1.0: From Images to High-Fidelity Simulation-Ready 3D Assets},
  author={Seed3D Team, ByteDance Seed},
  journal={Technique Report},
  year={2025}
}

@article{
oquab2024dinov,
title={{DINO}v2: Learning Robust Visual Features without Supervision},
author={Maxime Oquab and Timoth{\'e}e Darcet and Th{\'e}o Moutakanni and Huy V. Vo and Marc Szafraniec and Vasil Khalidov and Pierre Fernandez and Daniel HAZIZA and Francisco Massa and Alaaeldin El-Nouby and Mido Assran and Nicolas Ballas and Wojciech Galuba and Russell Howes and Po-Yao Huang and Shang-Wen Li and Ishan Misra and Michael Rabbat and Vasu Sharma and Gabriel Synnaeve and Hu Xu and Herve Jegou and Julien Mairal and Patrick Labatut and Armand Joulin and Piotr Bojanowski},
journal={Transactions on Machine Learning Research},
issn={2835-8856},
year={2024},
}

@article{ravi2024sam2,
  title={SAM 2: Segment Anything in Images and Videos},
  author={Ravi, Nikhila and Gabeur, Valentin and Hu, Yuan-Ting and Hu, Ronghang and Ryali, Chaitanya and Ma, Tengyu and Khedr, Haitham and R{\"a}dle, Roman and Rolland, Chloe and Gustafson, Laura and Mintun, Eric and Pan, Junting and Alwala, Kalyan Vasudev and Carion, Nicolas and Wu, Chao-Yuan and Girshick, Ross and Doll{\'a}r, Piotr and Feichtenhofer, Christoph},
  journal={arXiv preprint arXiv:2408.00714},
  year={2024}
}

@article{luo2023dhf,
  title={Diffusion Hyperfeatures: Searching Through Time and Space for Semantic Correspondence},
  author={Luo, Grace and Dunlap, Lisa and Park, Dong Huk and Holynski, Aleksander and Darrell, Trevor},
  journal={Advances in Neural Information Processing Systems},
  year={2023}
}

@inproceedings{radford2021learning,
  title={Learning transferable visual models from natural language supervision},
  author={Radford, Alec and Kim, Jong Wook and Hallacy, Chris and Ramesh, Aditya and Goh, Gabriel and Agarwal, Sandhini and Sastry, Girish and Askell, Amanda and Mishkin, Pamela and Clark, Jack and others},
  booktitle={International conference on machine learning},
  pages={8748--8763},
  year={2021},
  organization={PmLR}
}

@article{iliash2024s2o,
  title={{S2O}: Static to openable enhancement for articulated {3D} objects},
  author={Iliash, Denys and Jiang, Hanxiao and Zhang, Yiming and Savva, Manolis and Chang, Angel X},
  journal={arXiv preprint arXiv:2409.18896},
  year={2024}
}

@inproceedings{
wang2024pf,
title={{PF}-{LRM}: Pose-Free Large Reconstruction Model for Joint Pose and Shape Prediction},
author={Peng Wang and Hao Tan and Sai Bi and Yinghao Xu and Fujun Luan and Kalyan Sunkavalli and Wenping Wang and Zexiang Xu and Kai Zhang},
booktitle={The Twelfth International Conference on Learning Representations},
year={2024},
url={https://openreview.net/forum?id=noe76eRcPC}
}

@article{hong2023lrm,
  title={Lrm: Large reconstruction model for single image to 3d},
  author={Hong, Yicong and Zhang, Kai and Gu, Jiuxiang and Bi, Sai and Zhou, Yang and Liu, Difan and Liu, Feng and Sunkavalli, Kalyan and Bui, Trung and Tan, Hao},
  journal={arXiv preprint arXiv:2311.04400},
  year={2023}
}

@inproceedings{tang2024lgm,
  title={Lgm: Large multi-view gaussian model for high-resolution 3d content creation},
  author={Tang, Jiaxiang and Chen, Zhaoxi and Chen, Xiaokang and Wang, Tengfei and Zeng, Gang and Liu, Ziwei},
  booktitle={European Conference on Computer Vision},
  pages={1--18},
  year={2024},
  organization={Springer}
}

@inproceedings{szymanowicz2024splatter,
  title={Splatter image: Ultra-fast single-view 3d reconstruction},
  author={Szymanowicz, Stanislaw and Rupprecht, Chrisitian and Vedaldi, Andrea},
  booktitle={Proceedings of the IEEE/CVF Conference on Computer Vision and Pattern Recognition (CVPR)},
  pages={10208--10217},
  year={2024}
}

@article{TripoSR2024,
  title={TripoSR: Fast 3D Object Reconstruction from a Single Image},
  author={Tochilkin, Dmitry and Pankratz, David and Liu, Zexiang and Huang, Zixuan and Letts, Adam and Li, Yangguang and Liang, Ding and Laforte, Christian and Jampani, Varun and Cao, Yan-Pei},
  journal={arXiv preprint arXiv:2403.02151},
  year={2024}
}

@inproceedings{zhang2018perceptual,
  title={The Unreasonable Effectiveness of Deep Features as a Perceptual Metric},
  author={Zhang, Richard and Isola, Phillip and Efros, Alexei A and Shechtman, Eli and Wang, Oliver},
  booktitle={CVPR},
  year={2018}
}

\clearpage

\setcounter{topnumber}{4}
\setcounter{bottomnumber}{2}
\setcounter{totalnumber}{6}
\renewcommand{\topfraction}{0.95}
\renewcommand{\bottomfraction}{0.8}
\renewcommand{\textfraction}{0.05}
\renewcommand{\floatpagefraction}{0.5}
\renewcommand{\dbltopfraction}{0.95}
\renewcommand{\dblfloatpagefraction}{0.5}

\section{Additional Ablation Analyses}
\label{app:additional_ablations}

\begin{table*}[!t]
\centering
\small
\setlength{\tabcolsep}{6pt}
\renewcommand{\arraystretch}{1.15}
\caption{Additional ablations on PartNet-Mobility. The mask-dilation variant
tests sensitivity to imprecise mask boundaries, while the local-attention
variant tests the Stage~1 architectural design. The global-object-token
variant tests object-level denoising in Stage~2.}
\begin{tabular}{l|cccccccc}
\toprule
\textbf{Methods$\backslash$Metric} &
RS-$d_{\text{gIoU}}\downarrow$ & AS-$d_{\text{gIoU}}\downarrow$ &
RS-$d_{\text{cDist}}\downarrow$ & AS-$d_{\text{cDist}}\downarrow$ &
RS-$d_{\text{CD}}\downarrow$ & AS-$d_{\text{CD}}\downarrow$ &
AOR$\downarrow$ & CLIP$\uparrow$ \\
\midrule
\textbf{Ours} (GT Mask) & \textbf{0.1395} & \textbf{0.1695} &
\textbf{0.0202} & \textbf{0.0653} & \textbf{0.0016} &
\textbf{0.0139} & \textbf{0.0011} & \textbf{0.9053} \\
Ours (GT Mask + Random Dilation/Erosion) & 0.1491 & 0.1842 &
0.0264 & 0.0799 & 0.0016 & 0.0140 &
0.0043 & 0.9009 \\
Ours w/o Within-Part Local Attention & 0.2485 & 0.3317 &
0.0416 & 0.1285 & 0.0045 & 0.0267 &
0.0093 & 0.8932 \\
Ours w/o Global Object Token Sequence & 0.1655 & 0.1921 &
0.0285 & 0.0810 & 0.0016 & 0.0155 &
0.0097 & 0.8992 \\
\bottomrule
\end{tabular}
\label{tab:additional_ablations}
\end{table*}

\begin{figure*}[!t]
    \centering
    \includegraphics[width=0.98\textwidth]{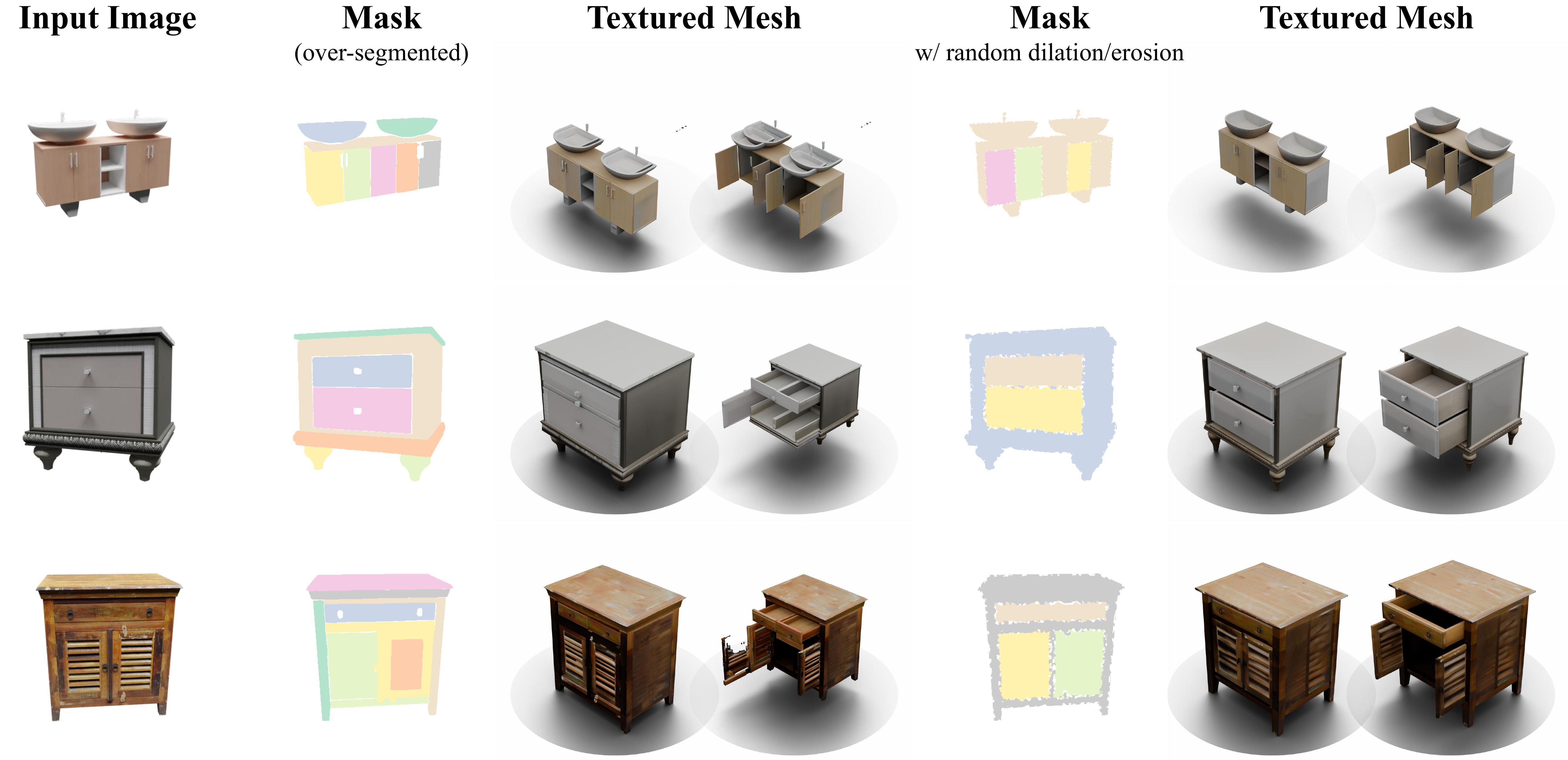}
    \caption{Qualitative results with imperfect masks. Each row shows the
    input image, a severely over-segmented mask and the resulting textured mesh,
    followed by a mask with random boundary dilation/erosion and its resulting
    textured mesh. \methodname tolerates moderate boundary perturbations, whereas
    severe part-instance segmentation errors remain challenging. Resting and
    articulated states are shown for each generated mesh.}
    \label{fig:mask_tolerance}
\end{figure*}

\paragraph{Sensitivity to mask quality.}
\methodname uses part masks as coarse localization cues rather than
pixel-accurate constraints. As shown in
Tab.~\ref{tab:additional_ablations}, random dilation or erosion of the GT
masks with a $5\times5$ kernel results in relatively limited degradation.
Figure~\ref{fig:mask_tolerance} further illustrates this distinction:
predictions remain broadly stable when the part identities are correct but
the mask boundaries contain moderate noise. In contrast, severely
over-segmented masks or incorrectly grouped mask regions alter the
part-instance specification provided to the model and can lead to structural
or articulation errors that the model cannot reliably correct. This also
helps explain the larger gap between the GT-mask oracle and the practical
VLM+SAM setting in Tab.~\ref{tab:metrics_pm}: performance depends more on
recovering the correct part instances than on obtaining pixel-perfect mask
boundaries.

\paragraph{Within-part local attention.}
Replacing within-part local attention with global attention at every layer would
also substantially increase computation and memory: for part-token lengths
$\{L_i\}_{i=1}^{K}$, a global layer scales quadratically with the total token
count, $\mathcal{O}((\sum_i L_i)^2)$, whereas a within-part layer costs
$\mathcal{O}(\sum_i L_i^2)$. Beyond this efficiency benefit, removing within-part
local attention degrades all reported motion-space metrics, including
RS-$d_{\mathrm{gIoU}}$ from 0.1395 to 0.2485 and AS-$d_{\mathrm{CD}}$ from
0.0139 to 0.0267 (Table~\ref{tab:additional_ablations}). This supports its role in preserving part-specific geometry,
while the interleaved global layers maintain cross-part context. PartCrafter~\cite{lin2025partcrafter}
reports a similar observation: removing its local-attention component causes
generation to collapse, supporting a hierarchical local--global design for
part-decomposed generation.

\paragraph{Global object token sequence in Stage~2.}
Our Stage~2 design inherits the object-level denoising formulation and pretrained
initialization of OmniPart~\cite{yang2025omnipart}. We concatenate the tokens from
all parts into a global object token sequence while preserving their identities
through part embeddings. This allows the pretrained denoiser to jointly model
cross-part spatial relationships and the overall object structure. To ablate this
design, we instead denoise each part-token sequence independently (``w/o Global Object Token Sequence'') while retaining the same joint-parameter post-processing. As shown in
Table~\ref{tab:additional_ablations}, this variant degrades all reported metrics except RS-$d_{\mathrm{CD}}$, which is unchanged,
including RS-$d_{\mathrm{gIoU}}$ from 0.1395 to 0.1655, AS-$d_{\mathrm{CD}}$
from 0.0139 to 0.0155, and AOR from 0.0011 to 0.0097. These results support
concatenating part tokens for globally coherent object-level denoising.

\section{Additional Implementation Details}
\label{sec:app_implementaion_details}

For SINGAPO~\cite{liu2024singapo}, a retrieval-based method, we use the official codebase to retrieve parts from its preprocessed datasets and provide the corresponding ground-truth articulation graphs as input. For ArtFormer~\cite{Su_2025_CVPR}, PhysX-3D~\cite{cao2025physx}, and PhysX-Anything~\cite{physxanything}, we conduct experiments using their official code and released checkpoints with their default inference settings unless stated otherwise.  For FreeArt3D~\cite{chen2025freeart3d}, we use the generated results released by the authors for comparison. Regarding Articulate-Anything~\cite{le2024articulate}, the official implementation depends on the deprecated Gemini-1.5-Flash API; therefore, we replace it with Gemini-2.5-Flash (a newer available endpoint) while keeping the original prompts and other steps unchanged. Below, we provide additional details on (i) the ArtFormer image-to-text surrogate setting, (ii) the CLIP-based similarity evaluation, and (iii) the PhysX-3D part-level articulation extraction used to match our evaluation protocol.

\begin{figure}[!t]
    \centering
    \includegraphics[width=0.98\linewidth]{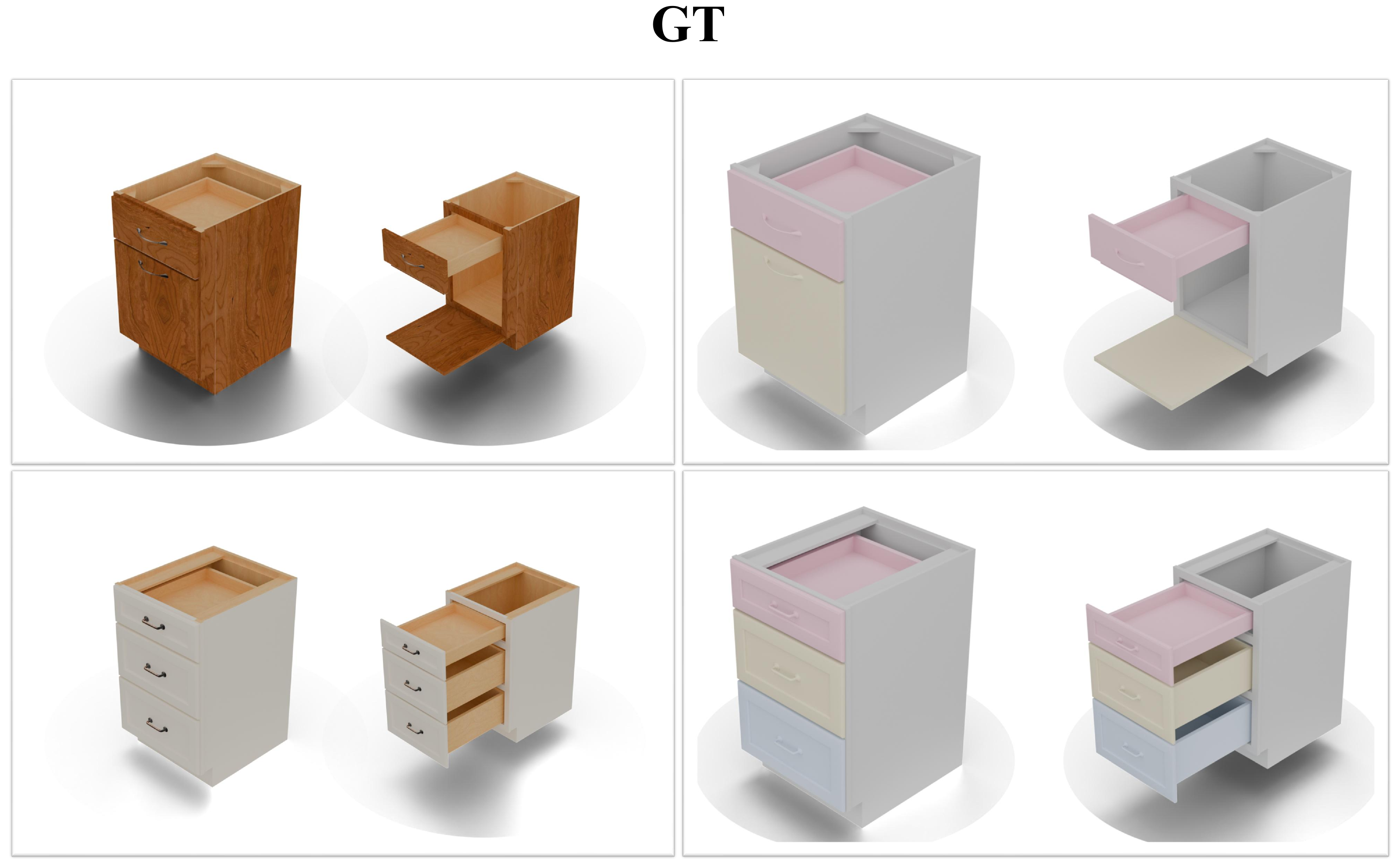}
    \vspace{-4mm}
    \caption{Ground-truth references for the FreeArt3D comparison in
    Fig.~\ref{fig:cmp_fa3d}. The left and right panels show textured and
    part-colored renderings, respectively, in both resting and articulated states.}
    \label{fig:cmp_fa3d_gt}
\end{figure}

\paragraph{ArtFormer Image-to-text Surrogate Setting.}
Since our benchmark provides images as inputs, we evaluate baselines under an image-conditioned setting when possible. However, ArtFormer does not release an image-conditioned model and does not provide the full training code to reproduce such a setting end-to-end. To enable a comparable image-based evaluation while staying faithful to their released \emph{text-based} model, we follow their preprocessing procedure to first generate an image description (caption) for each input image, converting the image condition into a textual prompt (\textit{image} $\rightarrow$ \textit{caption}). We then feed this text prompt into the official \textit{text-to-3D} generation pipeline. In other words, ArtFormer is evaluated through an ``\textit{image} $\rightarrow$ \textit{caption} $\rightarrow$ \textit{text-to-3D}'' surrogate pipeline that directly leverages its released text model.

\paragraph{CLIP-based Similarity Evaluation.}
For CLIP evaluation, we follow the protocol of FreeArt3D~\cite{chen2025freeart3d}. For each object, we sample $6$ articulation states, and for each state we render $6$ viewpoints. For each viewpoint, we render both the generated mesh and the corresponding ground-truth mesh under the same camera and compute CLIP \emph{image--image} similarity using CLIP \emph{ViT-L/14@336px}. We then average the similarities over all states and viewpoints to obtain a semantic similarity score per object. 
Since ArtFormer does not generate textures, we evaluate ArtFormer using renders of the \emph{untextured (white/gray) mesh} for \emph{both} its outputs and the corresponding ground-truth objects, to avoid bias due to missing textures; for all other methods we use textured renders.

\paragraph{PhysX-3D Articulation Extraction.}
PhysX-3D outputs a holistic object (e.g., a mesh) with per-vertex kinematic predictions, rather than explicit \emph{part-level} geometry and kinematic parameters. Therefore, we convert its per-vertex outputs into a part-level articulated representation to match our evaluation protocol. Specifically, we cluster vertices with the same predicted \emph{Part ID} to decompose the holistic shape into rigid parts. Meanwhile, we build the kinematic tree by assigning each part a \emph{Parent ID} via majority voting over its vertices, which improves robustness to boundary noise and defines the parent--child relations between parts. For joint parameters, we aggregate per-vertex predictions within each part: the joint type is determined by majority vote, while continuous attributes (axis direction, pivot point, and motion range) are averaged, with the axis direction re-normalized to obtain consistent kinematic parameters for each part.

\section{Post-processing of Predicted Articulation}
Because \methodname{} directly regresses the origin of each revolute joint,
small prediction errors may place the predicted hinge away from the associated
part. We therefore apply a deterministic refinement to revolute joints: the
regressed origin is replaced by its closest point on the surface of the part's
axis-aligned bounding box (AABB).

All main PAct results include this refinement, whereas baselines are reported
using their released inference procedures. To make the effect of this additional
step transparent, Table~\ref{tab:postprocess_ablation} also reports \methodname{}
without it under the same GT-mask setting. Without post-processing, \methodname{}
remains numerically better than every compared baseline across all seven metrics
in Table~\ref{tab:metrics_pm}. The projection improves five metrics, leaves
RS-$d_{\mathrm{CD}}$ unchanged, and changes AS-$d_{\mathrm{CD}}$ only slightly
from 0.0136 to 0.0139. Thus, the projection improves several reported metrics
but does not account for \methodname{}'s overall advantage.

\begin{table*}[!t]
\centering
\small
\caption{
Ablation of revolute-joint post-processing on PartNet-Mobility.
}
\label{tab:postprocess_ablation}
\begin{tabular}{lccccccc}
\toprule
Method & RS-$d_{\mathrm{gIoU}}\downarrow$ & AS-$d_{\mathrm{gIoU}}\downarrow$ &
RS-$d_{\mathrm{cDist}}\downarrow$ & AS-$d_{\mathrm{cDist}}\downarrow$ &
RS-$d_{\mathrm{CD}}\downarrow$ & AS-$d_{\mathrm{CD}}\downarrow$ & AOR$\downarrow$ \\
\midrule
Ours (with projection) & 0.1395 & 0.1695 & 0.0202 & 0.0653 &
0.0016 & 0.0139 & 0.0011 \\
Ours (without projection) & 0.1442 & 0.1941 & 0.0308 & 0.0774 &
0.0016 & 0.0136 & 0.0041 \\
\bottomrule
\end{tabular}
\end{table*}
\section{Extension to Complex Articulation Graphs}
\label{app:complex_graph}

Our current formulation follows prior articulated-object generation settings and focuses on depth-1 articulation trees, where each movable part is directly connected to a single fixed root. This setting covers many common articulated objects in embodied-AI scenarios, such as cabinets, drawers, doors, and appliances, but does not fully capture objects with deeper kinematic trees.

To study whether \methodname{} can be extended toward richer articulation structures, we conduct a preliminary experiment in which the Articulation Module is extended to additionally predict graph information, including parent-node indices. This variant, denoted as Ours (Graph Prediction), removes the fixed depth-1 topology assumption and attempts to infer parent-child relations from data.
As shown in Table~\ref{tab:graph_prediction}, explicitly predicting the articulation graph
does not improve the current model under the limited-data setting.
It leads to worse overlap-based metrics, including RS-$d_{\mathrm{gIoU}}$
and AS-$d_{\mathrm{gIoU}}$, as well as worse distance-based reconstruction and articulation metrics
such as RS-$d_{\mathrm{cDist}}$, AS-$d_{\mathrm{cDist}}$,
RS-$d_{\mathrm{CD}}$, and AS-$d_{\mathrm{CD}}$.
It also reduces CLIP similarity.
The graph-prediction variant can fit the training set but generalizes poorly to held-out objects.
These results suggest that \methodname{} is architecturally compatible with graph prediction,
but reliably learning general articulation trees remains challenging
under the current limited-data regime.

Therefore, we keep the depth-1 formulation in the main experiments for robustness and fair comparison with prior work. Extending \methodname{} to general depth-$N$ articulation trees with larger and more diverse articulated-object datasets remains an important direction for future work.

\begin{table*}[!t]
\centering
\small
\caption{
Ablation on extending the \textit{ArticulationModule} to predict articulation-graph information (PartNet-Mobility dataset). 
}
\vspace{-4mm}
\label{tab:graph_prediction}
\begin{tabular}{lcccccccc}
\toprule
\textbf{Methods$\backslash$Metric} & 
RS-$d_{\text{gIoU}}\downarrow$ & AS-$d_{\text{gIoU}}\downarrow$ & RS-$d_{\text{cDist}}\downarrow$ & AS-$d_{\text{cDist}}\downarrow$ & RS-$d_{\text{CD}}\downarrow$&
AS-$d_{\text{CD}}\downarrow$ & AOR$\downarrow$ & CLIP$\uparrow$ \\
\midrule
Ours (Graph Prediction) 
& 0.4114 & 0.4989 & 0.0964 & 0.1472 & 0.0018 & 0.0617 & 0.0811 & 0.8701 \\
Ours (w/o Graph Prediction) 
& 0.1395 & 0.1695 & 0.0202 & 0.0653 & 0.0016 & 0.0139 & 0.0011 & 0.9053 \\
\bottomrule
\end{tabular}
\end{table*}

\section{Comparison with PhysX-Anything}
\begin{figure}[!t]
    \centering
    \includegraphics[width=\linewidth]{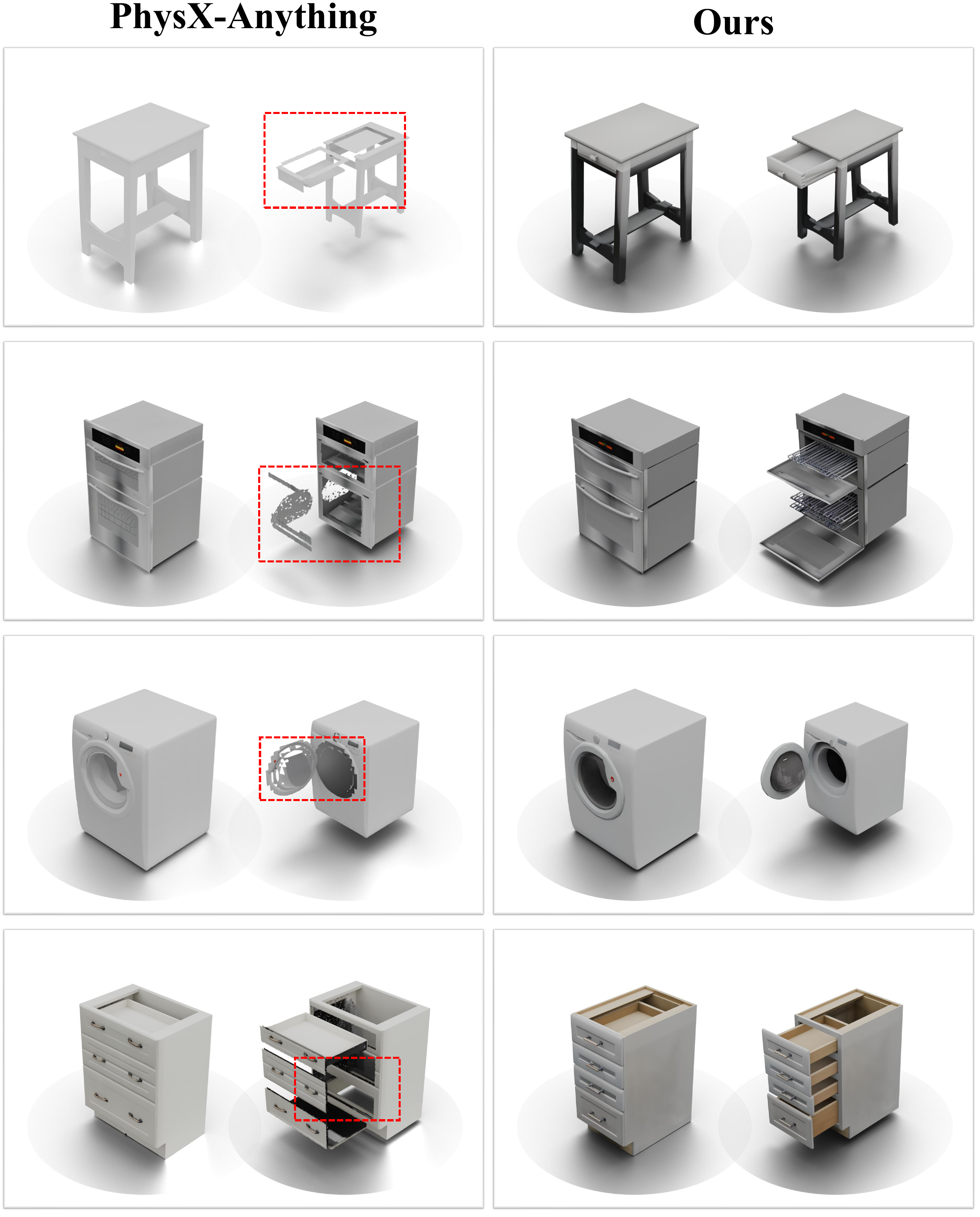}
    \vspace{-15pt}
    \caption{Qualitative comparison with the concurrent work PhysX-Anything~\cite{physxanything}.}
    \label{fig:cmp_physxanything}
\end{figure}

Tab. 1 (in the main text) and Tab. ~\ref{tab:metrics_acd} report quantitative comparisons with the concurrent work PhysX-Anything~\cite{physxanything}. To complement these results, we further provide a qualitative comparison in Fig.~\ref{fig:cmp_physxanything}. As highlighted by the red boxes, PhysX-Anything exhibits visible artifacts around movable parts, including incomplete geometry, holes, and noisy structures. In contrast, our method produces cleaner and more coherent part-level geometry, resulting in more plausible articulated objects under the same single-view setting.

\section{Qualitative Comparison with DIPO}
\label{app:dipo_qualitative}

\begin{figure*}[!t]
    \centering
    \includegraphics[width=0.80\textwidth]{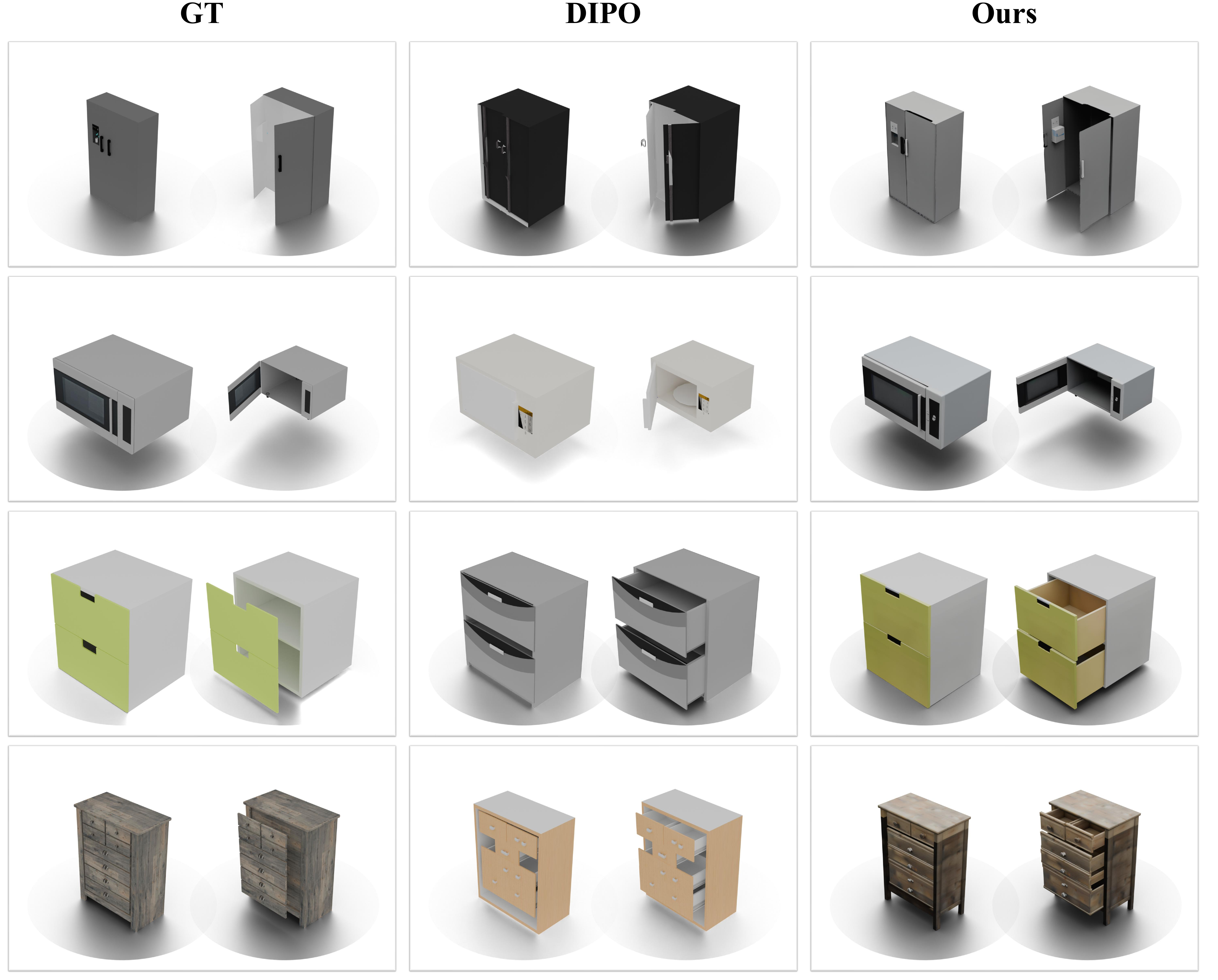}
    \vspace{-4mm}
    \caption{Qualitative comparison with DIPO~\cite{wu2025dipo}.
    Columns correspond to the ground truth (GT),
    DIPO, and \methodname; each cell shows the resting and articulated states.}
    \label{fig:cmp_dipo}
\end{figure*}

DIPO~\cite{wu2025dipo} is a strong quantitative baseline and uses images
from two articulation states, whereas \methodname uses a single image.
Figure~\ref{fig:cmp_dipo} provides the side-by-side comparison.
DIPO produces plausible category-level articulated objects through retrieval,
but its retrieved geometry and appearance can differ from the target instance.
Our generated results generally preserve more instance-specific appearance cues,
although neither method is free of articulation or reconstruction errors.

\section{Additional Real-World and Simulator Results}
\label{app:additional_applications}

\begin{figure*}[!t]
    \centering
    \includegraphics[width=0.9\textwidth]{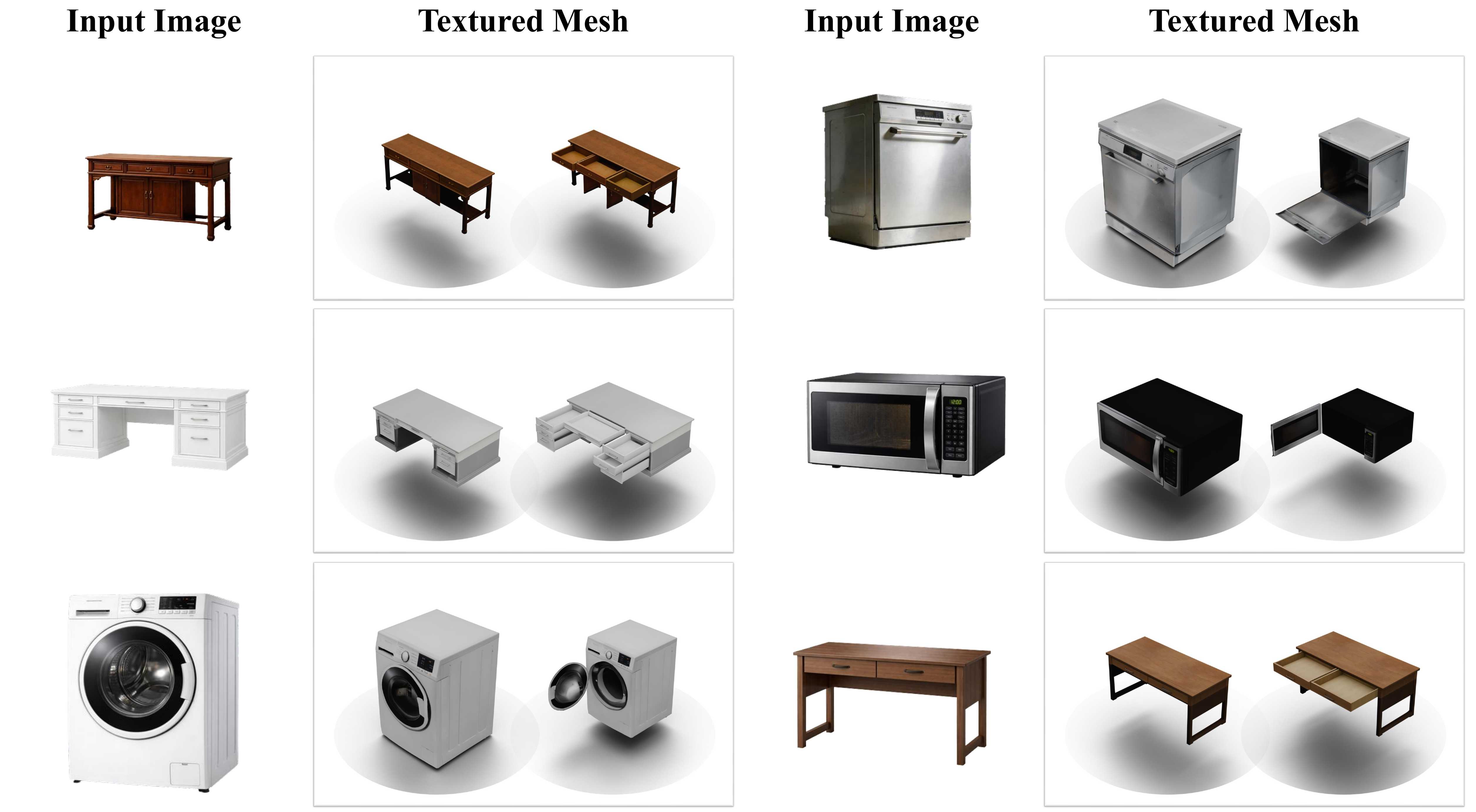}
    \vspace{-4mm}
    \caption{Additional qualitative results on in-the-wild inputs.
    For each example, the input image is followed by renders of the generated
    textured articulated mesh.}
    \label{fig:more_real}
\end{figure*}

Figure~\ref{fig:more_real} supplements the main-paper real-world results
with additional inputs and generated textured meshes. These examples are curated
and do not have corresponding ground-truth 3D assets; they therefore demonstrate
practical feasibility rather than providing a systematic robustness evaluation.

\begin{figure*}[!t]
    \centering
    \includegraphics[width=0.85\textwidth]{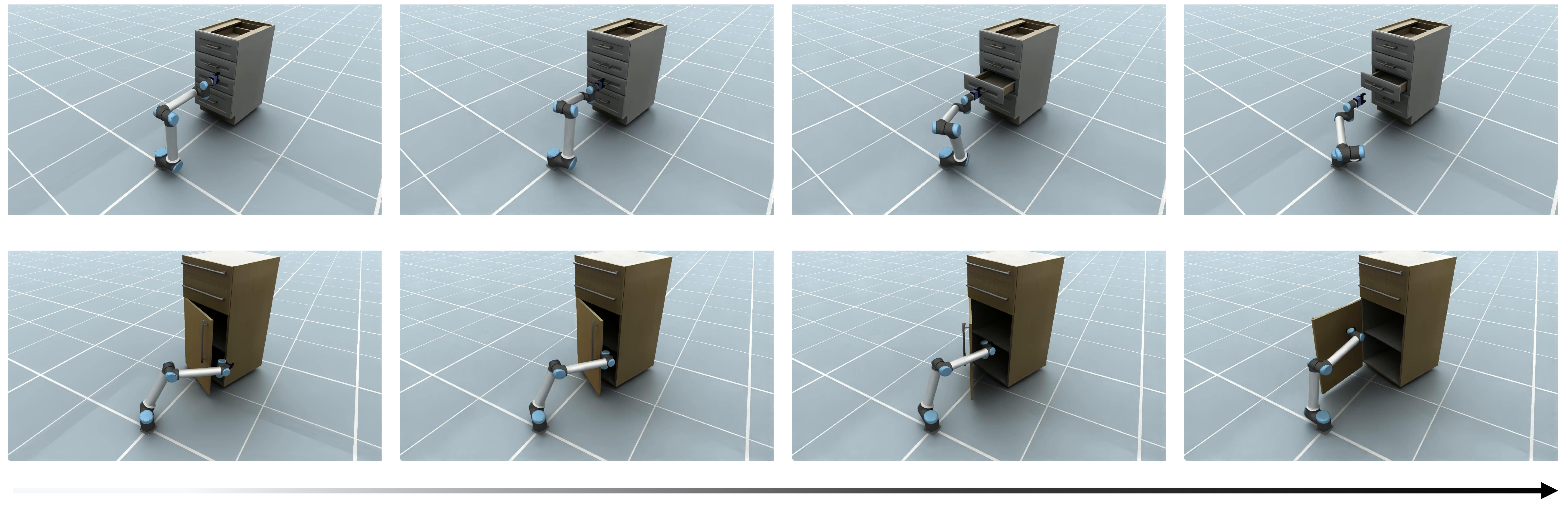}
    \vspace{-4mm}
    \caption{Qualitative simulator demonstration using generated assets.
    Consecutive frames from left to right show a robot interacting with generated
    drawers and a hinged door.}
    \label{fig:simulator}
\end{figure*}

We additionally import generated assets into a physics simulator and
interact with their predicted movable parts. Figure~\ref{fig:simulator}
shows representative manipulation sequences. This qualitative demonstration
checks that the generated representation can be instantiated and actuated in a
downstream simulator; it is not a large-scale robotics success-rate evaluation.

\section{Direct Evaluation of Joint Parameters}
\label{app:joint_param_eval}

In the main text, following prior generative articulated-object works~\cite{lei2023nap,liu2024singapo,wu2025dipo,liu2024cage}, our main evaluation adopts motion-space metrics rather than only reporting separate per-parameter errors. Specifically, geometry and articulation quality are evaluated in both the resting state (RS) and the articulated state (AS). This is because the final articulated motion is jointly determined by joint type, joint axis, pivot location, motion range, and part geometry. These parameters are non-linear and may not correspond equally to the final visual articulation state, making a uniform weighting of individual parameter errors less informative.

Nevertheless, we additionally report direct articulation-parameter metrics for reference. Specifically, we report joint-type accuracy, joint-axis error, and joint-pivot error on both PartNet-Mobility and ACD, following FreeArt3D~\cite{chen2025freeart3d}. As shown in
Table~\ref{tab:joint_metric}, \methodname{} obtains the highest joint-type accuracy
and lowest joint-pivot error on PartNet-Mobility,
but its joint-axis error (0.1065) is higher than DIPO's (0.0206).
DIPO uses two articulation states, whose motion cues directly facilitate axis estimation,
whereas \methodname uses a single image.
On the unseen ACD benchmark, \methodname{} obtains the lowest joint-axis and joint-pivot errors,
while its joint-type accuracy is below DIPO and SINGAPO.
These direct parameter results complement, but should not be conflated with,
the overall motion-space evaluation in the main text.

\begin{table}[!t]
\centering
\small
\caption{Direct evaluation of articulation parameters on
PartNet-Mobility (top) and ACD (bottom).}
\label{tab:joint_param_eval}

\label{tab:joint_metric}
\begin{tabular}{lccc}
\toprule
Method & Joint-Type-Acc $\uparrow$ & Joint-Axis-Err $\downarrow$ & Joint-Pivot-Err $\downarrow$ \\
\midrule
SINGAPO & 0.9408 & 0.1357 & 0.0370 \\
DIPO & 0.9599 & 0.0206 & 0.1561 \\
ArtFormer & 0.3835 & 0.4187	& 0.2358 \\
Arti-Anything & 0.8853 & 0.2094 & 0.3351 \\
PhysX-3D & 0.4360 & 0.9696 & 0.3229 \\
PhysX-Anything & 0.8020 & 0.3591 & 0.4628 \\
Ours  & 0.9675 & 0.1065 & 0.0219 \\
\bottomrule
\end{tabular}

\begin{tabular}{lccc}
\toprule
Method & Joint-Type-Acc $\uparrow$ & Joint-Axis-Err $\downarrow$ & Joint-Pivot-Err $\downarrow$ \\
\midrule
SINGAPO & 0.8322 &0.3136  &	0.2506 \\
DIPO & 0.8735 & 0.2149 & 0.2128 \\
ArtFormer & 0.3745  & 0.9429	 &0.2945 \\
Arti-Anything & 0.4848 & 1.2807 & 0.3789 \\

PhysX-3D & 0.3102 &0.9664 &0.3017 \\
PhysX-Anything & 0.5588 & 0.6724 & 0.2475 \\
Ours  & 0.8142 & 0.1350 & 0.0956 \\
\bottomrule
\end{tabular}

\end{table}

\section{Appearance Fidelity Evaluation}
\label{sec:appearance_evaluation}

The qualitative comparison in Fig.~\ref{fig:cmp_baseline} and the failure
examples in Fig.~\ref{fig:failure_physical_appearance} show that the generated
appearance may deviate from the reference, particularly in hollow or occluded regions
where texture supervision is ambiguous. To complement this qualitative analysis, we
quantitatively evaluate the rendered appearance fidelity of the generated articulated
objects. For each object, we match the generated and ground-truth renderings by object
identity and articulation state, yielding five matched rendering pairs per object.

To prevent the transparent background from dominating the evaluation, we
compute the union of the foreground alpha masks for each matched pair and use
its tight bounding box as a shared crop. We apply the same crop to the generated
and ground-truth renderings, resize both cropped images to the same resolution,
and composite them onto a white background. This shared preprocessing preserves
their relative scale and alignment.

We report three complementary perceptual metrics. LPIPS~$\downarrow$ ~\cite{zhang2018perceptual}
measures perceptual dissimilarity using deep image features. DINOv2
similarity~$\uparrow$ measures feature-level visual correspondence using the
cosine similarity between image embeddings extracted by DINOv2~\cite{oquab2024dinov}
with the ViT-L/14-reg backbone.
CLIP image similarity (CLIP)~$\uparrow$ measures higher-level semantic and
appearance correspondence using the cosine similarity between image embeddings
extracted by CLIP~\cite{radford2021learning} with the ViT-L/14 backbone. These
metrics complement, rather
than replace, the geometry and articulation metrics reported in the main paper.
For each dataset and metric, we first average the five articulation-state scores
for each object and then average the resulting object-level scores over the
dataset. Thus, every object has equal weight within its dataset.

\begin{table}[!t]
    \centering
    \caption{
        Appearance fidelity of our generated articulated objects with respect
        to the corresponding ground-truth renderings. Each dataset score averages
        object-level values, with each object evaluated under five articulation
        states. Higher DINOv2 and CLIP similarities and lower LPIPS indicate
        better correspondence.
    }
    \label{tab:appearance_fidelity}
    \setlength{\tabcolsep}{6pt}
    \begin{tabular}{lccc}
        \toprule
        Dataset
        & DINOv2 Sim.$\uparrow$
        & LPIPS$\downarrow$
        & CLIP$\uparrow$ \\
        \midrule
        PartNet-Mobility  & 0.8129 & 0.1921 & 0.9148 \\
        ACD & 0.6363 & 0.2489 & 0.8654 \\
        \bottomrule
    \end{tabular}
\end{table}

Table~\ref{tab:appearance_fidelity} shows the same cross-dataset trend for
all three metrics: compared with ACD, PM has higher DINOv2 and CLIP
similarities and lower LPIPS. Under these metrics, the generated objects
therefore preserve the appearance of their ground-truth counterparts more
consistently on PM than on ACD.

\section{Capability Comparison with Existing Methods}
In Table~\ref{tab:ability}, we compare representative feedforward  methods and summarize their key capabilities. \textit{Geometry} indicates whether the method synthesizes
instance-specific 3D shape (\eg, meshes, point clouds, or Gaussian-based
representations) rather than reusing retrieved part meshes. \textit{Texture} indicates whether it synthesizes or reconstructs
appearance rather than inheriting it from retrieved assets.  \textit{Part-level Articulation} indicates whether the method
outputs semantic part-level geometry together with articulation parameters
that can be directly used to animate the generated object.  \textit{Non-Retrieval} indicates whether the method operates without retrieving parts or examples
from a predefined asset database at inference time.

\begin{table*}[!t]
\centering
\small
\setlength{\tabcolsep}{6pt}
\renewcommand{\arraystretch}{1.15}
\caption{Comparison with baselines on the ACD Dataset~\cite{iliash2024s2o}.}
\vspace{-4mm}
\begin{tabular}{l|cccccccc}
\toprule
\textbf{Methods$\backslash$Metric} & 
RS-$d_{\text{gIoU}}\downarrow$ & AS-$d_{\text{gIoU}}\downarrow$ & RS-$d_{\text{cDist}}\downarrow$ & AS-$d_{\text{cDist}}\downarrow$ & RS-$d_{\text{CD}}\downarrow$&
AS-$d_{\text{CD}}\downarrow$ & AOR$\downarrow$ & CLIP$\uparrow$ \\
\midrule
SINGAPO~\cite{liu2024singapo} & 0.7272 & 0.7306 & 0.1585 & 0.1944 & 0.0090 & 0.0256 & 0.0115 & 0.8376 \\
DIPO~\cite{wu2025dipo} & 0.9927 & 0.9940 & 0.1379 & 0.1747 & 0.0081 & 0.0198 & 0.0200 & 0.8517 \\

ArtFormer~\cite{Su_2025_CVPR} & 1.3489 & 1.3876 & 0.3456 & 0.4414 & 0.0396 & 0.0825 & 0.0147 & 0.7690 \\
PhysX-3D~\cite{cao2025physx} & 0.6449 & 0.6464 & 0.2591 & 0.3396 & 0.0209 & 0.0677 & 0.0960 & 0.8132 \\

PhysX-Anything~\cite{physxanything} & 1.1673 & 1.1709 & 0.2965 & 0.3950 & 0.0162 & 0.0481 & \textbf{0.0099} & 0.8456 \\
Arti-Anything~\cite{le2024articulate} & 1.2677 & 1.2693 & 0.5632 & 0.5781 & 0.2583 & 0.2803 & 0.1236 & 0.8648 \\
\midrule
Ours (VLM+SAM Mask) & 0.5316 & 0.5412 & 0.1249 & 0.1656 & 0.0065 & 0.0218 & 0.0248 & \textbf{0.8901} \\
\midrule
\textbf{Ours} (GT Mask) & \textbf{0.4182} & \textbf{0.4289} & \textbf{0.0830} & \textbf{0.1142} & \textbf{0.0060} & \textbf{0.0194} & 0.0107 & 0.8862 \\
\bottomrule
\end{tabular}
\label{tab:metrics_acd}
\end{table*}

\begin{table*}[!t]
\centering
\small
\setlength{\tabcolsep}{15pt} 
\renewcommand{\arraystretch}{1.15}
\caption{Comparison of representative methods and their capabilities.}
\vspace{-3mm} 
\begin{tabular}{l ccc c}
\toprule
Methods &  Geometry & Texture & Part-level Articulation & Non-Retrieval\\
\midrule
SINGAPO~\cite{liu2024singapo}  & $\times$ & $\times$ & \ding{51} & $\times$\\
DIPO~\cite{wu2025dipo}  & $\times$ & $\times$ & \ding{51} & $\times$\\
ArtFormer~\cite{Su_2025_CVPR}  & \ding{51} & $\times$ & \ding{51}  & \ding{51} \\
PhysX-3D~\cite{cao2025physx}  & \ding{51} & \ding{51} & $\times$ & \ding{51} \\
Articulate-Anything~\cite{le2024articulate}   & $\times$ & $\times$ & \ding{51} & $\times$\\
\midrule
\textbf{Ours}  & \ding{51} & \ding{51} & \ding{51} & \ding{51}\\
\bottomrule
\end{tabular}
\label{tab:ability}
\end{table*}

\section{VLM-Guided Part Analysis and Mask Annotation}
\label{app:vlm_mask}

Articulated objects often admit multiple plausible part decompositions due to the presence of several movable components. To improve the controllability of articulated part generation, we incorporate an additional 2D part mask $\mathcal{M}$ as a conditioning input, which explicitly specifies the desired part decomposition in the input image.

To automate mask acquisition, we first use SAM2~\cite{ravi2024sam2} to produce fine-grained segmentation masks. Since SAM2 may over-segment an articulated part into multiple fragments, we further employ a vision--language model (VLM, \textit{e.g.}, GPT-5) to identify masks that belong to the same part and merge them into a single part mask. This produces a cleaner, part-level decomposition that better matches the articulation structure and serves as our final conditioning mask $\mathcal{M}$. The entire pipeline is built upon the OmniPart~\cite{yang2025omnipart} codebase and consists of the following stages.

\begin{figure}[!t]
    \centering
    \includegraphics[width=0.9\linewidth]{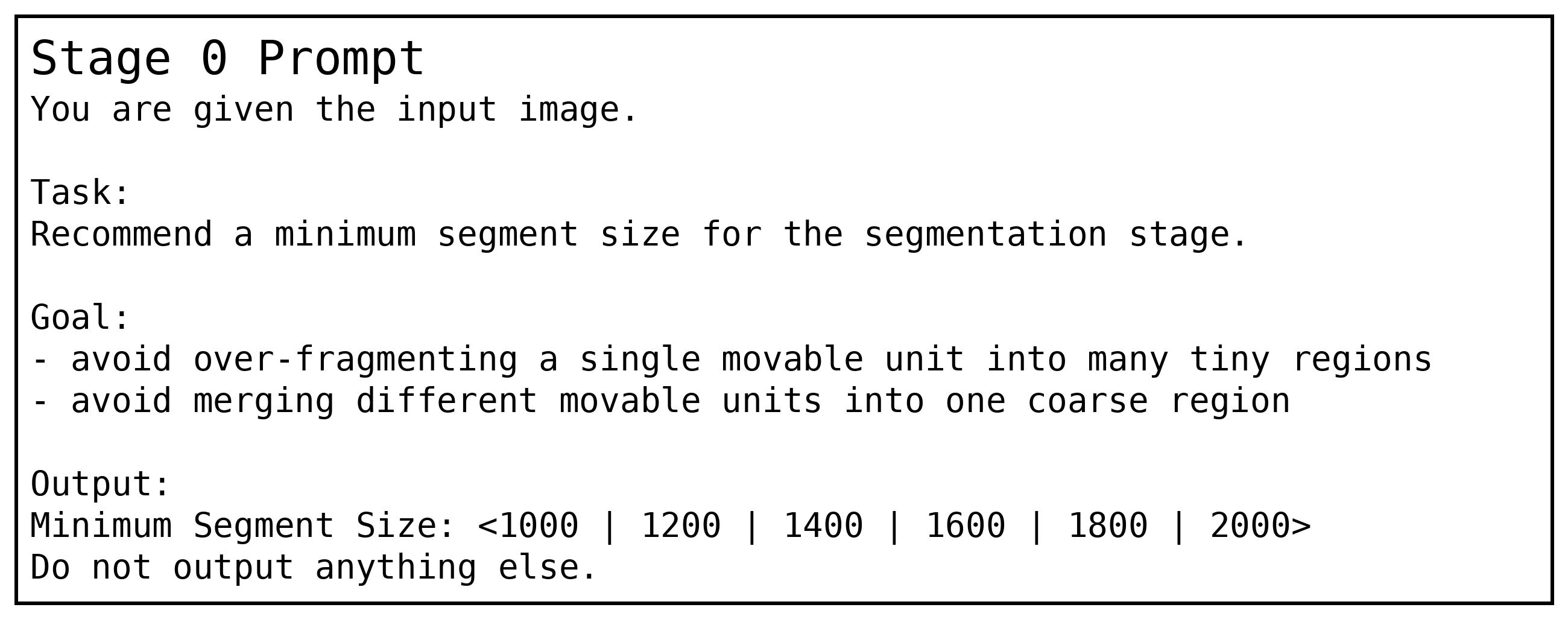}
    \caption{Prompt used in Stage 0 for segmentation granularity selection.}
    \label{fig:gpt5_prompt_0}
\end{figure}

\paragraph{Stage 0: Pre-Process}
Prior to segmentation, we employ GPT-5 (\texttt{gpt-5-2025-08-07}) to determine the granularity for part segmentation, utilizing the prompting strategy illustrated in Fig.~\ref{fig:gpt5_prompt_0}. More specifically, the model takes as input the original image together with an explicit instruction to predict a discrete minimum segment size, and outputs one value from a predefined candidate set.

\begin{figure}[!t]
    \centering
    \includegraphics[width=0.9\linewidth]{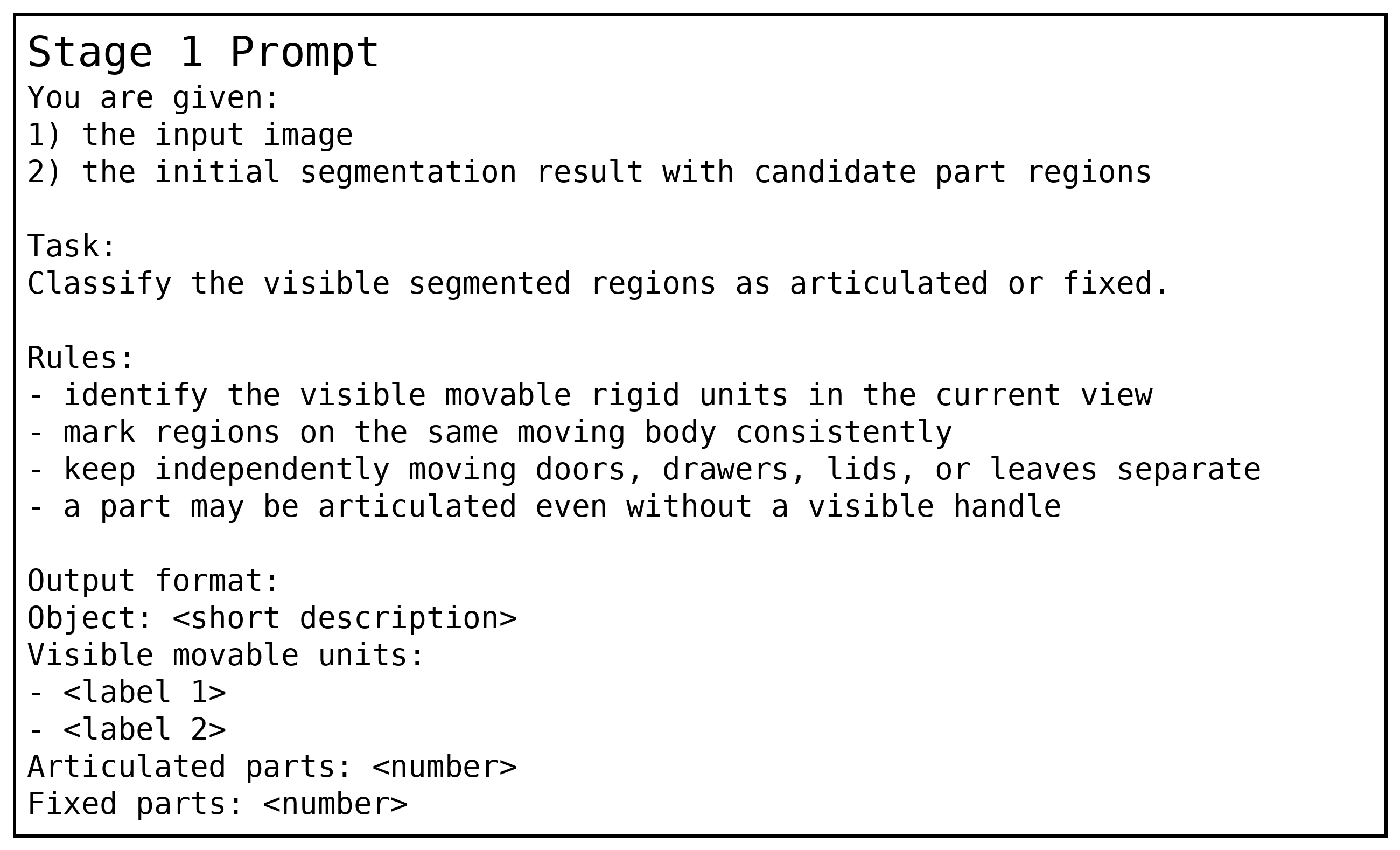}
    \caption{Prompt used in Stage 1 for visible part classification.}
    \label{fig:gpt5_prompt_1}
\end{figure}

\paragraph{Stage 1: Part Classification}
In the classification phase, the Segment Anything Model (SAM)~\cite{ravi2024sam2} first generates candidate part segmentations with the granularity proposed in the pre-processing stage. The VLM is then prompted, as shown in Fig.~\ref{fig:gpt5_prompt_1}, to classify each segment as either articulated or fixed. Given the original image and the initial segmentation result, the model identifies the visible movable rigid units and predicts which segmented regions belong to articulated components and which belong to fixed structure. 

\begin{figure*}[!t]
    \centering
    \begin{minipage}[t]{0.47\textwidth}
        \centering
        \includegraphics[width=\linewidth]{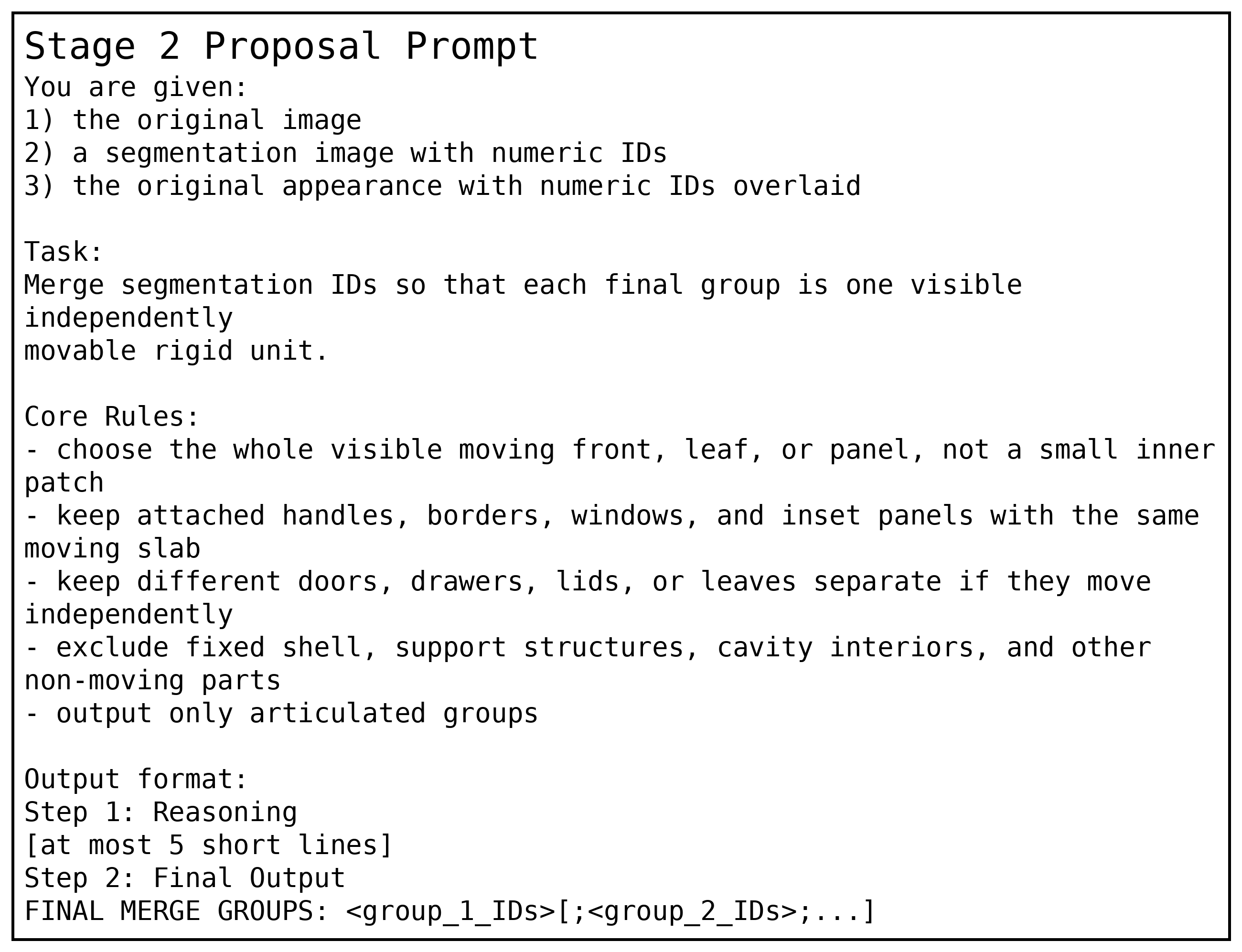}
        \vspace{-5mm}
        \caption*{(a) Stage 2 proposal prompt.}
    \end{minipage}
    \hfill
    \begin{minipage}[t]{0.47\textwidth}
        \centering
        \includegraphics[width=\linewidth]{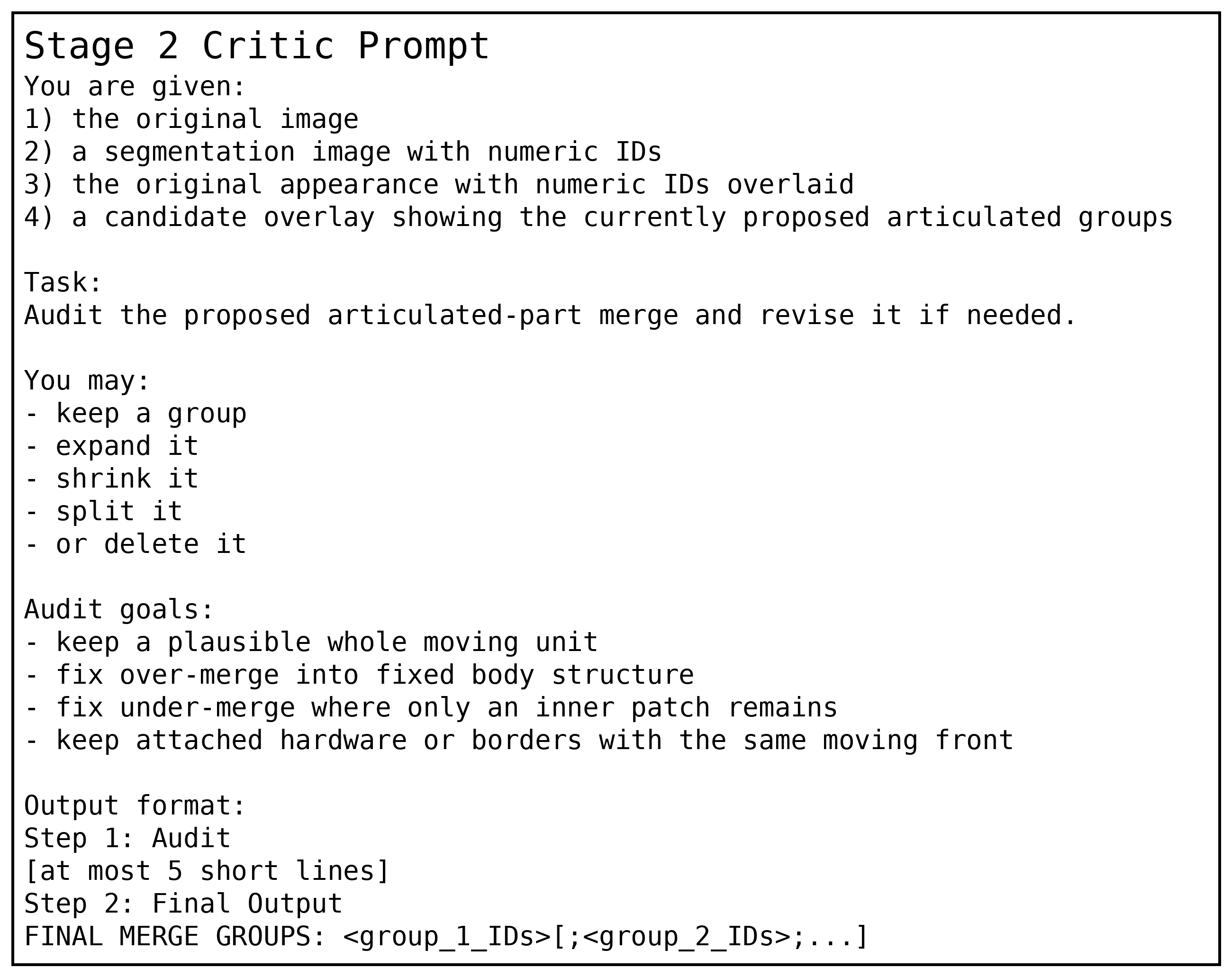}
        \vspace{-5mm}
        \caption*{(b) Stage 2 critic prompt.}
    \end{minipage}
    \vspace{-3mm}
    \caption{Prompts used in Stage 2 for articulated-part merging.}
    \label{fig:gpt5_prompt_2}
\end{figure*}

\paragraph{Stage 2: Part Merging}
Building upon the labeled RGB images and part classifications obtained in Stage 1, the VLM performs a semantic merging process to consolidate segments that belong to the same articulated part. For example, SAM2~\cite{ravi2024sam2} often treats a door and its handle as two distinct entities. In our setting, we require these components to be recognized as a single kinematic body. Concretely, this stage is implemented as a proposal-critic pipeline. In the proposal step, the VLM predicts initial merge groups from the original image, the numbered segmentation map, and the numbered overlay on the original image. In the critic step, the VLM further audits and revises the proposal using an additional candidate overlay that visualizes the currently predicted articulated groups. The proposal and critic prompts used in this stage are shown in Fig.~\ref{fig:gpt5_prompt_2}. This merging process effectively corrects the over-segmentation artifacts produced by SAM2, yielding a coherent, articulation-aware segmentation that serves as our final conditioning input $\mathcal{M}$.
The final articulated groups are taken from the critic output and then rendered back into a part-level segmentation mask, which is used as the conditioning input $\mathcal{M}$.

\section{Per-Category Evaluation on PartNet-Mobility}
Table~\ref{tab:per_category_metrics} provides a fine-grained per-category comparison across different articulated object classes. Overall, our method achieves consistently strong performance across most categories, especially in geometry-related metrics such as RS-$d_{\text{CD}}$ and AS-$d_{\text{CD}}$. Compared with prior methods, the improvements are particularly clear on Oven and Table, where our method obtains the lowest Chamfer distances and joint-pivot errors. On Microwave and WashingMachine, our method achieves the lowest RS-$d_{\text{CD}}$, but DIPO attains lower AS-$d_{\text{CD}}$, and on WashingMachine Articulate-Anything obtains better bounding-box metrics. For larger and more diverse categories such as StorageFurniture, our method also maintains competitive performance, demonstrating robustness to intra-category structural variation. We observe that Dishwasher remains relatively challenging, mainly due to larger geometric variation and more ambiguous part decomposition from single-view inputs.

\begin{table*}[t]
\centering
\caption{Per-category quantitative comparison on PartNet-Mobility dataset.}
\label{tab:per_category_metrics}
\resizebox{\textwidth}{!}{
\begin{tabular}{llcccccccccc}
\toprule
Method & Category & 
RS-$d_{\text{gIoU}}\downarrow$ & AS-$d_{\text{gIoU}}\downarrow$ & RS-$d_{\text{cDist}}\downarrow$ & AS-$d_{\text{cDist}}\downarrow$ & RS-$d_{\text{CD}}\downarrow$&
AS-$d_{\text{CD}}\downarrow$ & AOR$\downarrow$ &  Joint-Type-Acc $\uparrow$ & Joint-Axis-Err$\downarrow$ & Joint-Pivot-Err $\downarrow$  \\
\midrule

\multirow{7}{*}{SINGAPO}
& Dishwasher  & 0.5972 & 0.6101 & 0.1031 & 0.3125 & 0.0052 & 0.1404 & -- & 1.0000 & 0.0039 & 0.1492 \\
& Microwave  & 0.3856 & 0.3759 & 0.0534 & 0.0312 & 0.0053 & 0.0056 & -- & 1.0000 & 0.0025 & 0.0191 \\
& Oven  & 0.5222 & 0.5341 & 0.1034 & 0.1209 & 0.0042 & 0.0130 & 0.0000 & 0.7500 & 0.3427 & 0.0511 \\
& Refrigerator & 0.8410 & 0.8416 & 0.1747 & 0.1271 & 0.0029 & 0.0039 & 0.0086 & 1.0000 & 0.0016 & 0.0235 \\
& StorageFurniture  & 0.4582 & 0.4725 & 0.0918 & 0.1293 & 0.0027 & 0.0209 & 0.0112 & 0.9462 & 0.1692 & 0.0285 \\
& Table & 0.5179 & 0.5373 & 0.0795 & 0.1179 & 0.0093 & 0.0151 & 0.0000 & 0.9200 & 0.0899 & 0.0077 \\
& WashingMachine  & 0.3355 & 0.3522 & 0.0898 & 0.0429 & 0.0067 & 0.0062 & 0.0001 & 1.0000 & 0.0279 & 0.0375 \\

\midrule
\multirow{7}{*}{ArtFormer}
& Dishwasher & 1.2789 & 1.2865 & 0.3834 & 0.6759 & 0.0552 & 0.3121 & 0.0081 & 0.5000 & 1.4738 & 0.3072 \\
& Microwave & 1.2937 & 1.3009 & 0.3555 & 0.7150 & 0.0371 & 0.3796 & 0.0089 & 0.0000 & 0.0294 & 0.6919 \\
& Oven & 1.3290 & 1.3336 & 0.4506 & 0.6043 & 0.0510 & 0.1433 & 0.0027 & 0.3300 & 1.2614 & -- \\
& Refrigerator & 1.3013 & 1.3090 & 0.2829 & 0.4494 & 0.0337 & 0.1441 & 0.0126 & 0.2917 & 0.1569 & 0.2178 \\
& StorageFurniture & 1.3381 & 1.3422 & 0.3526 & 0.5046 & 0.0507 & 0.1919 & 0.0222 & 0.4179 & 0.1539 & 0.1937 \\
& Table & 1.2343 & 1.2389 & 0.3518 & 0.4570 & 0.0533 & 0.1091 & 0.0154 & 0.2850 & 1.2969 & 0.2241 \\
& WashingMachine & 1.2502 & 1.2590 & 0.4337 & 0.4947 & 0.0658 & 0.1285 & 0.0105 & 0.0000 & 0.6735 & 0.6498 \\

\midrule
\multirow{7}{*}{Arti-Anything}
& Dishwasher & 0.5990 & 0.6074 & 0.1010 & 0.5012 & 0.0180 & 0.2666 & -- & 0.8500 & 0.0000 & 0.3068 \\
& Microwave & 1.0920 & 1.0908 & 0.3056 & 0.6119 & 0.0232 & 0.2588 & 0.0000 & 1.0000 & 0.0000 & 0.2730 \\
& Oven & 0.1299 & 0.1551 & 0.0091 & 0.2613 & 0.0036 & 0.0508 & 0.0000 & 0.9000 & 0.1257 & 0.3546 \\
& Refrigerator & 0.9695 & 0.9777 & 0.2144 & 0.3327 & 0.0076 & 0.0713 & 0.0061 & 1.0000 & 0.0000 & 0.2197 \\
& StorageFurniture & 0.6735 & 0.6837 & 0.1330 & 0.3352 & 0.0107 & 0.1236 & 0.0069 & 0.8674 & 0.2717 & 0.3483 \\
& Table & 0.8613 & 0.8702 & 0.1941 & 0.2861 & 0.0061 & 0.0396 & 0.0000 & 0.9600 & 0.0000 & -- \\
& WashingMachine & 0.0613 & 0.1000 & 0.0043 & 0.1018 & 0.0067 & 0.0265 & 0.0000 & 1.0000 & 0.0000 & 0.3071 \\

\midrule
\multirow{7}{*}{DIPO}
& Dishwasher & 0.6646 & 0.6543 & 0.1021 & 0.2807 & 0.0036 & 0.1278 & -- & 0.9000 & 0.0037 & 0.2735 \\
& Microwave & 0.4948 & 0.5144 & 0.0447 & 0.0398 & 0.0031 & 0.0029 & -- & 1.0000 & 0.0005 & 0.0177 \\
& Oven & 0.6359 & 0.6512 & 0.0453 & 0.1186 & 0.0031 & 0.0083 & 0.0289 & 0.8333 & 0.2931 & 0.2549 \\
& Refrigerator & 0.8075 & 0.8090 & 0.0797 & 0.1312 & 0.0030 & 0.0154 & 0.0024 & 0.9500 & 0.0023 & 0.1390 \\
& StorageFurniture & 0.5836 & 0.5902 & 0.0504 & 0.0924 & 0.0020 & 0.0135 & 0.0657 & 0.9726 & 0.0482 & 0.1503 \\
& Table & 0.6351 & 0.6567 & 0.0463 & 0.0928 & 0.0066 & 0.0131 & 0.0000 & 0.9800 & 0.0013 & 0.1176 \\
& WashingMachine & 0.4472 & 0.4440 & 0.0210 & 0.0224 & 0.0017 & 0.0017 & 0.0000 & 1.0000 & 0.0317 & 0.1017 \\

\midrule
\multirow{7}{*}{PhysX-3D}
& Dishwasher & 0.7292 & 0.7279 & 0.3974 & 0.5728 & 0.0104 & 0.2515 & -- & 0.6250 & 1.0591 & 0.2457 \\
& Microwave & 0.9620 & 0.9705 & 0.3301 & 0.6109 & 0.0091 & 0.3041 & -- & 0.5000 & 0.9512 & 0.3434 \\
& Oven & 0.8827 & 0.8848 & 0.3324 & 0.4655 & 0.0095 & 0.0643 & -- & 0.3330 & 0.9570 & 0.1556 \\
& Refrigerator & 0.8906 & 0.8992 & 0.3309 & 0.4566 & 0.0056 & 0.1115 & -- & 0.5833 & 1.0867 & 0.4117 \\
& StorageFurniture & 0.8064 & 0.8094 & 0.3374 & 0.4793 & 0.0098 & 0.1228 & 0.2125 & 0.4288 & 0.9538 & 0.3396 \\
& Table & 0.4243 & 0.4244 & 0.1967 & 0.2575 & 0.0244 & 0.0606 & -- & 0.3650 & 0.9672 & 0.2051 \\
& WashingMachine & 0.8400 & 0.8400 & 0.4753 & 0.5507 & 0.0097 & 0.0329 & -- & 0.3333 & 1.0670 & 0.0770 \\

\midrule
\multirow{7}{*}{PhysX-Anything}
& Dishwasher & 1.0457 & 1.0551 & 0.2446 & 0.5251 & 0.0067 & 0.1725 & 0.0038 & 0.7500 & 1.5708 & 0.3901 \\
& Microwave & 0.9878 & 0.9949 & 0.1026 & 0.5251 & 0.0061 & 0.2191 & -- & 1.0000 & 0.0000 & 0.3995 \\
& Oven & 1.0854 & 1.0881 & 0.2851 & 0.4101 & 0.0100 & 0.0700 & 0.0015 & 0.7833 & 1.1074 & 0.3480 \\
& Refrigerator & 0.8489 & 0.8526 & 0.2491 & 0.3329 & 0.0071 & 0.0407 & -- & 0.8667 & 0.3927 & 0.2793 \\
& StorageFurniture & 0.9928 & 1.0010 & 0.2551 & 0.4029 & 0.0101 & 0.1309 & 0.0071 & 0.8528 & 0.2948 & 0.5022 \\
& Table & 0.9130 & 0.9187 & 0.2790 & 0.3950 & 0.0112 & 0.0532 & 0.0042 & 0.5150 & 0.0000 & 0.1152 \\
& WashingMachine & 0.8276 & 0.8369 & 0.1034 & 0.1641 & 0.0073 & 0.0371 & 0.0062 & 0.6667 & 0.0008 & 0.2858 \\

\midrule
\multirow{7}{*}{Ours}
& Dishwasher & 0.3277 & 0.3357 & 0.1263 & 0.2120 & 0.0030 & 0.0599 & -- & 1.0000 & 0.1963 & 0.1299 \\
& Microwave & 0.0212 & 0.0685 & 0.0036 & 0.0341 & 0.0019 & 0.0033 & -- & 1.0000 & 0.0000 & 0.0000 \\
& Oven & 0.0886 & 0.1129 & 0.0064 & 0.0289 & 0.0016 & 0.0024 & 0.0063 & 1.0000 & 0.0066 & 0.0000 \\
& Refrigerator & 0.2562 & 0.2743 & 0.0314 & 0.0592 & 0.0039 & 0.0082 & 0.0008 & 1.0000 & 0.0982 & 0.0030 \\
& StorageFurniture & 0.1253 & 0.1564 & 0.0146 & 0.0603 & 0.0014 & 0.0141 & 0.0006 & 0.9545 & 0.1267 & 0.0169 \\
& Table & 0.1140 & 0.1531 & 0.0096 & 0.0493 & 0.0009 & 0.0026 & 0.0031 & 1.0000 & 0.0090 & 0.0000 \\
& WashingMachine & 0.1771 & 0.1822 & 0.0091 & 0.0380 & 0.0015 & 0.0030 & 0.0000 & 1.0000 & 0.0000 & 0.0000 \\

\bottomrule
\end{tabular}
}
\label{tab:cmp_per_cat}
\end{table*}

\end{document}